\documentclass[preprint,review]{elsarticle}

\usepackage{hyperref}
\usepackage[singlespacing]{setspace}
\usepackage{enumitem}
\usepackage{amsmath} 
\usepackage{amsfonts}
\usepackage{dirtytalk}
\usepackage{xcolor}
\usepackage{multicol}
\usepackage{booktabs}
\usepackage{algorithm}
\usepackage{algpseudocode}
\usepackage{longtable}
\usepackage{eurosym}
\usepackage{multirow}
\usepackage{array}
\usepackage{subfig}

\newcommand{\nop}[1]{}

\definecolor{bleudefrance}{rgb}{0.19, 0.55, 0.91}
\journal{}

\begin{document}

\begin{frontmatter}

\title{Parcel loss prediction in last-mile delivery: deep and non-deep approaches with insights from Explainable AI}


\author{Jan de Leeuw\corref{cor2}} 
\author{Zaharah Bukhsh\corref{cor2}} 
\author{Yingqian Zhang \corref{cor1}} 
\ead{yqzhang@tue.nl}
\cortext[cor1]{Corresponding author}
\address{Department of Industrial Engineering \& Innovation Sciences, Eindhoven University of Technology, P.O. Box 513,
5600 MB Eindhoven, The Netherlands}






\begin{abstract}
Within the domain of e-commerce retail, an important objective is the reduction of parcel loss during the last-mile delivery phase. The ever-increasing availability of data, including product, customer, and order information, has made it possible for the application of machine learning in parcel loss prediction. However, a significant challenge arises from the inherent imbalance in the data, i.e., only a very low percentage of parcels are lost.  In this paper, we propose two machine learning approaches, namely, Data Balance with Supervised Learning (DBSL) and Deep Hybrid Ensemble Learning (DHEL), to accurately predict parcel loss. The practical implication of such predictions is their value in aiding e-commerce retailers in optimizing insurance-related decision-making policies.  
We conduct a comprehensive evaluation of the proposed machine learning models using one year data from Belgian shipments. The findings show that the DHEL model, which combines a feed-forward autoencoder with a random forest, achieves the highest classification performance.  Furthermore, we use the techniques from Explainable AI (XAI) to illustrate how prediction models can be used in enhancing business processes and augmenting the overall value proposition for e-commerce retailers in the last mile delivery.  
\end{abstract}


\end{frontmatter}


\section{Introduction}


The e-commerce sector plays a vital role in today's global economy. In 2020, e-commerce accounted for 15.5\% of the global retail sales \citep{Statista2020} and was projected to reach 22\% by the end of 2023. This growth is driven by technological innovation and shifting consumer behavior patterns \citep{clement2019commerce}. Data-driven approaches like machine learning allow retailers to gain insights into customer purchasing trends, optimize pricing and demand forecasting, and enhance pre- and post-purchase customer service. Products sold on e-commerce platforms require physical delivery, which 
is considered as one of the fundamental factors determining consumers' decision in selecting e-commerce retailers \citep{vakulenko2019online}. 

The last-mile delivery is defined as the final step in the supply chain as the parcel is shipped from the last distribution center to the end customer or collection point \citep{gevaers2011characteristics}. 
The last-mile delivery has has emerged as a major cost center for e-commerce retailers, accounting for 30\% of the total costs and 53\% of total shipping costs \citep{dolan2021challenges}.  The 
ability to meet the anticipated delivery dates and quantities \citep{bopage2019strategic} is directly tied to customer satisfaction \citep{suguna2021study}.  Failure to fulfill orders as promised can negatively impact the customer experience and brand perception. Therefore, optimizing last-mile logistics through improved planning and execution is strategically important. 
Unfortunately, parcel loss is often observed in last-mile delivery. To reduce the costs associated with lost shipments, especially for high-value items, many retailers opt to purchase insurance through their external logistics partners. These agreements typically stipulate that the delivery company will reimburse the economic damages from any parcels lost during last-mile transportation.  
However, selectively insuring only high-risk orders remains challenging without a systematic way to identify shipments more prone to getting lost.

The current approaches in parcel loss prediction tend to be non-data-driven and rule-based. The ever-increasing availability of data has made it
possible for the application of machine learning (ML) in parcel loss prediction. 
Machine learning applications in similar research domains are sparse, although the potential is high. 
However, the parcel loss problem suffers from the class imbalance problem, as only a small percentage of delivered parcels is lost. For example, our industry partner observed only 0,25\% of deliveries resulted in losses.  This brings a challenge to developing accurate machine learning models, especially in predicting the minority class \citep{xu2006classification}.  
In this work, we develop machine learning prediction models that are accurate, fast and generally applicable for current and future delivery service companies related to the last-mile delivery. Furthermore, model development and analyses seek to increase understanding of key drivers behind lost parcels. These actionable insights have the potential to enable strategic, data-driven improvements in operational planning and execution by retailers 
subsequently leading to a reduction in the incidence of parcel losses. 

Our work contributes to both theory and practice in the following ways: 
From a theoretical perspective, we contribute to the nascent literature at the intersection of machine learning and last-mile delivery logistics by:
\begin{itemize}
    \item Providing the first application of machine learning for predictive analytics of parcel loss risks within last-mile delivery logistics.
    \item Introducing two machine learning frameworks - DBSL and DHEL - designed specifically to address the challenges posed by highly imbalanced non-time series input data related to customer, order, and product attributes. The proposed deep hybrid ensemble learning method (DHEL) takes inspiration from deep anomaly detection techniques, transcending their conventional use in time series or image-centric data domains. The DHEL detects, on average 55,4\% of all lost parcels, with an average balanced accuracy 0,701, which is more accurate than the current business rule (42,2\% and 0,561, respectively).
\end{itemize}

Practically, we aim to benefit e-commerce retailers and delivery service providers by:
\begin{itemize}
    \item Enabling more accurate identification of at-risk shipments to optimize insurance spending and reduce costs. Through our case study with a large e-commerce retailer, we demonstrate how leveraging machine learning techniques can enrich business process insights and drive enhanced business value through cost optimization. Specifically, with our proposed DBSL model, the company that we did a case study, could save upto \euro550.600,80 annually in insurance premium costs by selectively targeting only high-risk deliveries.
    \item Informing delivery process improvements to lower incident rates by identifying key predictive attributes such as customer, product, and contextual attributes associated with parcel losses. We leverage explainable AI techniques to reveal the relative importance of features.  We also provide an elaborate overview of the tradeoff between prediction performance and interpretability of two proposed machine learning models. This deliberative analysis serves as a valuable guide for making informed decisions regarding the future adoption of these models within the e-commerce retail domain.
\end{itemize}

The rest of this paper is structured as follows. We discuss the related work of parcel loss prediction and classification models with highly imbalanced data in \ref{sec: Literature Review}. 
We then present a case study in Section \ref{sec:usecase}. 
Thereafter, the available data is explored in Section \ref{sec:data} and data insights are extracted and discussed. We introduce our modeling approaches to predict parcel loss in Section \ref{sec:methods}. In Section \ref{sec:performance}, we discuss the results of the predictive models. Subsequently, we present the business impact and additional process insights gained from prediction models in Section \ref{sec:business}. %
Section \ref{sec:conclude} concludes our study.

\section{Literature Review}
\label{sec: Literature Review}
%
We first describe the parcel loss problem. Then, we discuss the applications of machine learning to related problems in logistics.  Subsequently, we introduce existing techniques to handle highly imbalanced data. 
\paragraph{Parcel loss} 
The last-mile delivery starts at the distribution depot. Parcels are sorted and loaded into transportation vehicles, driving a predefined route of stops to deliver the loaded parcels. The logistics service modes for last-mile delivery can be distinguished into two categories, i.e., direct delivery mode and indirect delivery mode \citep{li2020logistics}. In direct delivery, the parcel is directly delivered to the customer and handed to the customer face to face. 
This mode has the advantage of high safety, reliability, and customer satisfaction. However, it has relatively low efficiency and a high failure rate of first-time delivery. 
Moreover, with unattended deliver, i.e., delivery that is left unsecured at home, theft is the main problem and a form of fraud. 
\cite{mckinnon2003unattended} study the relationship between unattended delivery and fraud, and demonstrate that crime risk in America depends on the size, value, and degree of concealment of the parcel and the nature of the neighborhood. 
A recent study by \cite{stickle2020porch} states that visibility from the roadway and easily recognisable brands or other indicators of high-value parcels are critical identifiers for parcel fraud. 
In the indirect delivery mode, parcels are delivered to self-pick-up locations and the customer has to pick up the parcel from this location. This mode improves efficiency by reducing failure rates of first-time delivery, but comes at the cost of reduced customer satisfaction and potential fraud risk \citep{savelsbergh201650th}. 
Researchers have also investigated different parcel tracking technologies over the past years by using RFID, barcodes, QR codes, or GPS. Systems automatically and continuously collect tracking data, such as the parcel's location, which can help decrease the amount of parcel loss \citep{ma2018design}. 
New methods, such as drones, autonomous vehicles, and trunk delivery, have been extensively studied, e.g., \citep{khoufi2019survey, rojas2021unmanned}.  
\cite{chen2021blockchain} propose a blockchain-based intelligent anti-switch parcel tracing logistics system. 
In this system, logistics operators can continuously monitor the status query of parcels by attaching a sensor to the parcel. 
The abovementioned new technologies show promising opportunities for reduction of parcel loss, but currently lack reliability. 

\paragraph{Machine learning in logistics} 
In the literature, machine learning has shown its promising performance in several applications within last mile logistics. 
For instance, the authors of \citep{pegado2023data} use classical machine learning techniques, such as tree-based models and linear regression models, to predict how many delivery and pickup services will remain uncompleted on a given route with the working day of a courier. 
With a case study, they show the proposed approach achieves promising results. 
The studies of \citep{de2019deep} and \citep{Gmira-2020} show that the deep learning models are better than non-deep ML models (i.e., Random Forests, XGBoost, and Support Vector Regression) 
in the task of estimating origin-destination travel time in parcel deliveries.
\cite{mo2023predicting} propose a pair-wise attention-based pointer neural network to predict route trajectories, using drivers’ historical delivery trajectory data. 
\cite{mathew2021defraudnet} present an end-to-end framework for detecting fraudulent transactions in food delivery based on self-generated weak labels. The proposed model is an ensemble of an autoencoder and an LSTM, using the reconstruction error of the autoencoder as input to a Multilayer Perceptron. 
\cite{lorenc2018intelligent} aim to predict the probability of cargo theft in road transport by using archive information about transportation, transport theft, type of transport, and transport value. 
In another related application area, i.e., baggage loss prediction at airports, 
\cite{van2020lost} propose a gradient boosting machine to identify bags at risk, showing more accurate prediction compared to conventional decision rule methods. 


To conclude, in the literature, some interesting relationships are observed between features and the amount of parcel loss from the literature, such as the size, stock value, and delivery method. However, 
there are currently no methods to do quantitative prediction of parcel loss prior to delivery. Yet, machine learning has been identified as useful techniques in other applications within last-mile logistics. 
In addition, for predicting parcel loss, an additional challenge is that data is highly imbalanced, i.e., only very few parcels are lost. 
This makes it hard to directly apply the standard ML approaches.  

\paragraph{Anomaly detection for imbalanced data}
Anomaly detection techniques can tackle the class imbalance problem by treating the minority class instances as outliers. 
Deep learning (DL) for anomaly detection utilizes neural networks, which have shown advantages over traditional algorithms \citep{chalapathy2019deep}. DL approaches are often well-adapted to jointly model the interactions between multiple variables beyond the specification of hyperparameters. Therefore, deep anomaly detection models require minimal tuning to obtain adequate results. Deep learning models can model complex, nonlinear relationships within the data. Moreover, in deep anomaly detection no feature engineering is required and the performance can scale with the availability of training data, making deep models suitable for data-rich problems \citep{choi2021deep}. Semi-supervised deep anomaly detection techniques are widely adopted in the literature, because normal instances are easier to obtain. Semi-supervised techniques use existing labels of normal instances to separate them from outliers. Semi-supervised deep anomaly detection techniques use all training instances from the normal class to learn a discriminative boundary around these normal instances \citep{song2017hybrid}. Test instances that do not belong to the normal class are then labeled as anomalous \citep{perera2019learning}. Autoencoders are the most widely used approach for performing semi-supervised deep anomaly detection. Training of the autoencoder is done in a semi-supervised manner. The autoencoder is first fed with normal data, meaning only the majority class is presented to the autoencoder, which it learns to reproduce. Afterwards, the validation set containing both labeled classes, is used for supervised parameter tuning by setting a reconstruction error threshold, distinguishing anomalies from the normal behavior. Semi-supervised learning can produce considerably improved performance over unsupervised techniques \citep{chalapathy2019deep}.

Autoencoders 
have shown great potential for classification on highly skewed data. However, most handle 
either time-series or image input data. 
Hence, it remains unknown what the performance of deep anomaly detection techniques is on non-time series, tabular, data for parcel loss prediction. 

\section{Case Study}
\label{sec:usecase}
The case study was conducted with a large Dutch e-commerce retailer (referred to as \emph{Company X}). 
Company X 
operates both online and through over ten physical retail stores located across countries in BeNeLux. 
Each day, thousands of online ordered products are picked from its warehouse and shipped to hubs, stores, and customers. Company X promises a guaranteed next-day delivery, seven days a week. To do so, it uses its own deliver network, which we name as OwnDN, and two external parcel delivery companies, named as ExtD1 and ExtD2, respectively, for the last-mile delivery, where 
more than 70 percent of all parcels are delivered via external delivery companies. 
Based on data in 2020, on average, 0,33\% of all shipped parcels are lost in the process from the warehouse to the customer. There are various reasons for parcel loss, varying from process errors to fraud. However, Company X  can only globally trace back the parcel, namely until the last received scan event. This last scan determines the phase in the supply chain where the parcel got lost. 
After analysis of parcel loss data in 2020, 
we notice that almost half of all parcels (around 45\%) are lost at the customer, and around 15\% of the lost parcels occur at the driver and around 15\% at the depot. This indicates that around 75\% of all parcel loss occurs in the last-mile delivery. In addition, the high percentage of loss at the customer and driver sparkles the idea that most parcel loss is due to fraudulent behavior of humans instead of process errors.

\begin{figure}[hb]
    \centering
    \includegraphics[width = 0.65\textwidth]{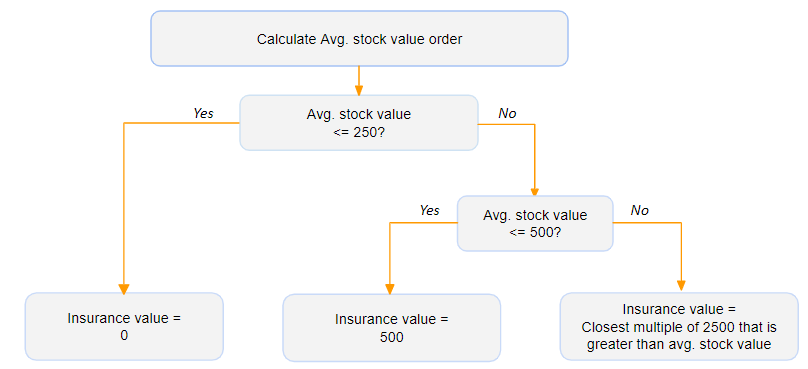}
    \caption{Insurance rules used by an external delivery partner (ExtD2) of Company X}
    \label{fig: bpost }
\end{figure}

Company X attempts to reduce the costs of parcel loss in the last-mile delivery by using insurance on high-valued parcels delivered by external partners. 
The insurance is an agreement between Company X and the delivery partner, which states that the delivery partner covers the economic damage of parcel loss in the last-mile delivery. 
In the current situation, domain experts have established insurance rules based on stock value and
product category. 
The rules determine each parcel's insurance value, which is the maximum amount of money Company X gets back from the delivery partner in case of parcel loss. We illustrate the insurance rule used for one deliver partner in Figure \ref{fig: bpost }. 
It can be observed that the current insurance value is only based on the stock value of the order. Orders with a stock value below \euro250 are not insured. 
Given the insurance value of a parcel, the insurance cost is then determined. 
For confidentiality reasons the exact costs are not shown. 
The insurance costs
differ per delivery partner as the delivery partners differ in the
amount of parcel loss. 
Company X is particularly interested in high predictive performance on the minority class, i.e., prediction of lost parcels, while remaining accurate predictions on the majority class. Therefore, ML models should be able to overcome the data-imbalance problem. In the current situation, Company X is inconclusive on individual parcel loss causes. 
Hence, understanding what feature attributes are related to parcel loss could be used to further optimize the last-mile delivery. 

\section{Data Description and Pre-processing}
\label{sec:data}


The collected data consists of shipments in Belgium in 2020, with products having a stock value above \euro100. 
The initial data contains about one million shipped products.   
The data contain order, customer, and product related attributes (or features).  
Order related data describes information about the carrier, delivery method, delivery address, route, etc. 
Customer related features contains the information of the customers who placed an order. This information is privacy sensitive, and hence is not used in further analysis and modelling.  
Product related features include size, weight, and stock value of each product. 

\begin{figure}[tb]
    \subfloat[Numerical and boolean features\label{fig: corrmatrix 1}]{%
      \includegraphics[width=0.5\textwidth]{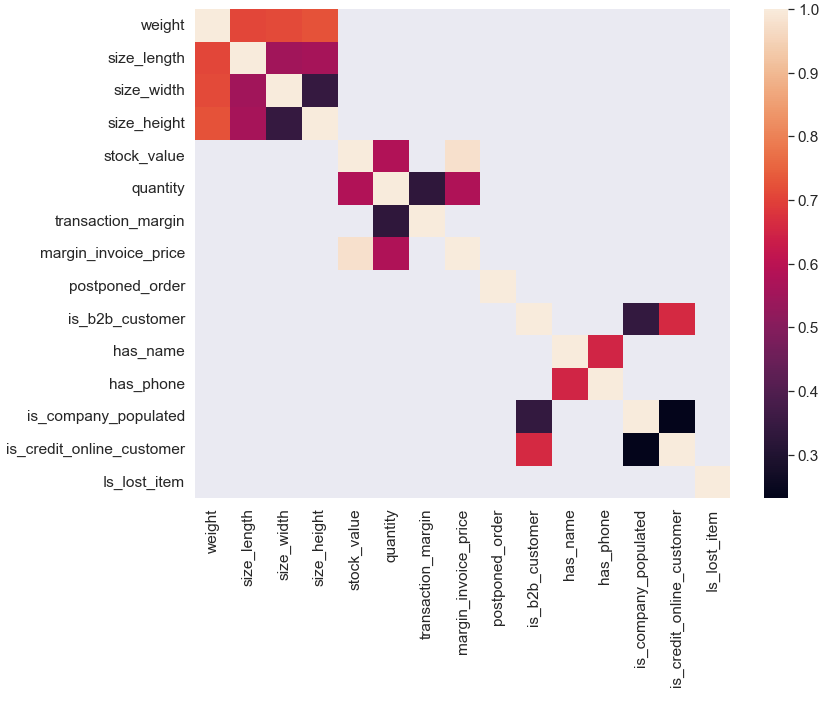}
    }
    \hfill
    \subfloat[Categorical features \label{fig: corrmatrix 2}]{%
      \includegraphics[width=0.5\textwidth]{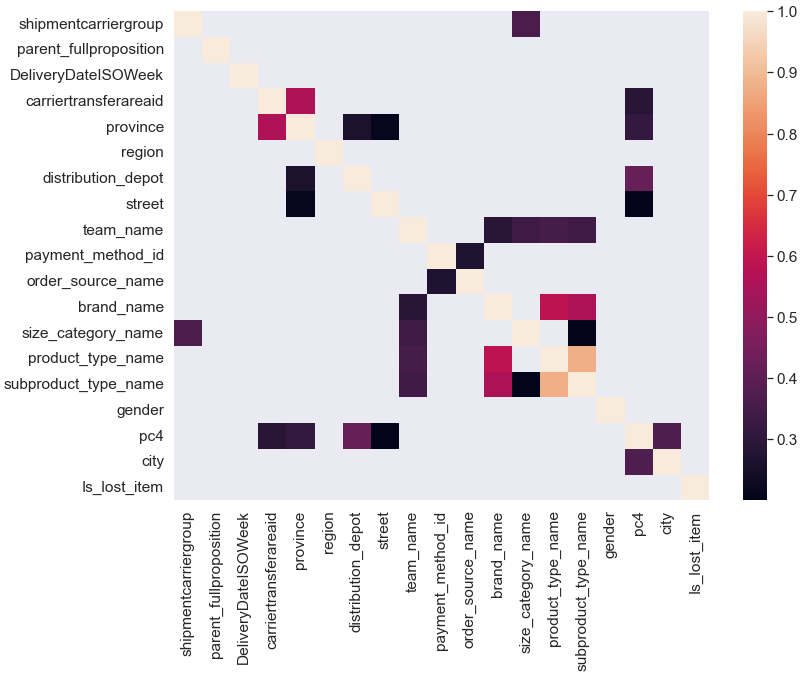}
    }
    \caption{Correlation matrix of features with target variable}
\end{figure}

To predict whether an order is lost, we create the target variable, \textit{Is\_lost\_item}, which is true (1) if the product is lost in the last-mile delivery, and otherwise false (0). 
We observe the class imbalance problem in our use case, as the distribution of the created binary target variable is highly imbalanced, i.e., only 0,54\% of products are lost, and 99,46\% are normal parcel deliveries. 

Figure \ref{fig: corrmatrix 1} shows the correlation matrix of the numerical and boolean features. 
We notice that no feature is observed to have a weak linear correlation with the target feature \textit{Is\_lost\_item}. Moreover, there exist some correlations between predictive numerical and boolean features. It can be observed that the \textit{size} and \textit{weight} features are moderately correlating. Moreover, the \textit{price-related} features are weakly correlated. Similarly, in Figure \ref{fig: corrmatrix 2}, it shows 
there are weak and moderate correlations between predictive features, such as \textit{shipment} features. The \textit{address} features are weakly correlating, namely street, pc4, distribution\_depot, and city. Moreover, some \textit{product} properties, such as product type and brand, are moderately correlated. Similar to the numerical features, no linear correlation between the features and the target variable is observed. This suggests linear models might not be able to provide accurate predictions. 

\begin{figure}[tb]
    \subfloat[Boxplot of size\_length\label{fig: Pareto}]{%
      \includegraphics[width=0.45\textwidth]{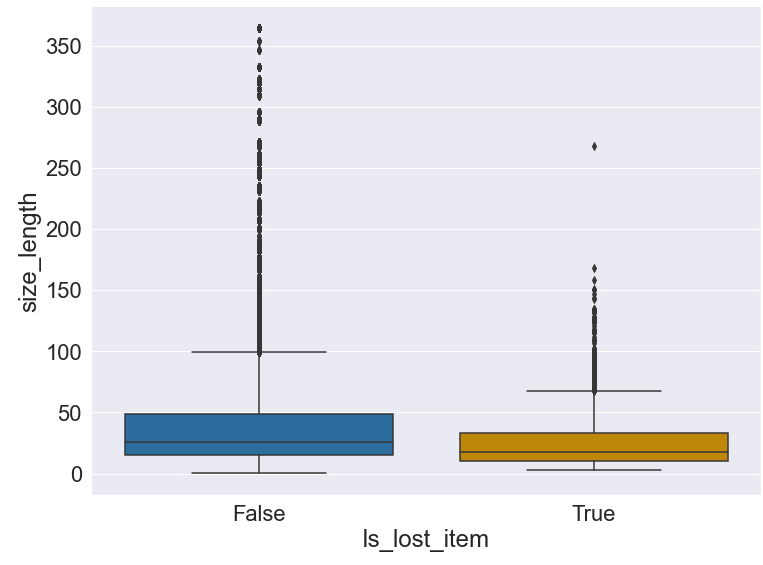}
    }
    \hfill
    \subfloat[Distribution is\_b2b\_customer\label{fig: Pareto Close up}]{%
      \includegraphics[width=0.5\textwidth]{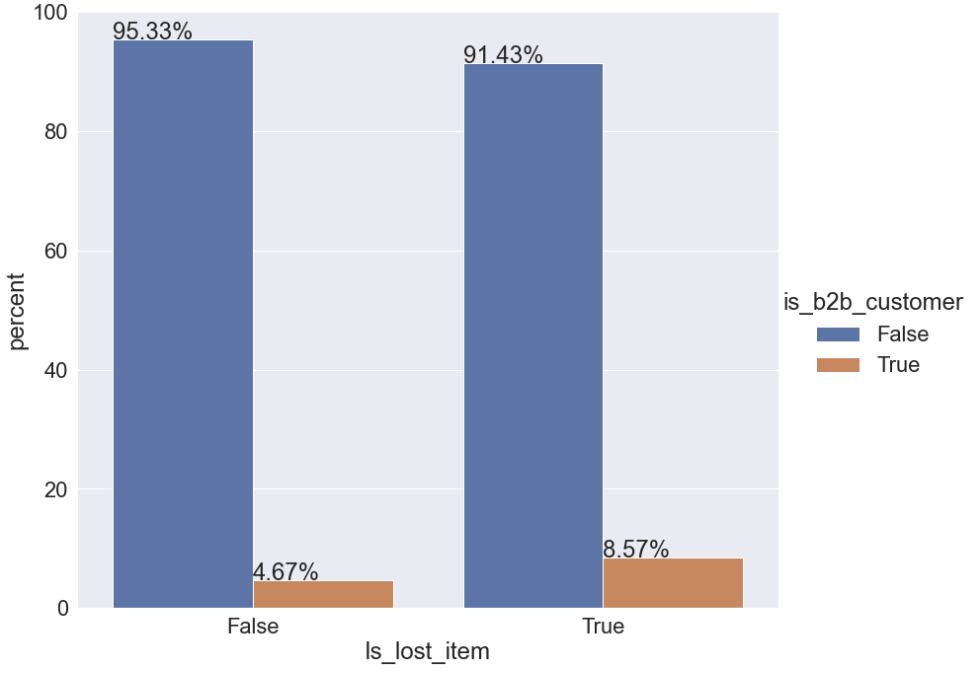}
    }
    \caption{Distribution of predictors LI's vs. not LI's}
    \label{fig: BuddyProgramm plot}
\end{figure}

We perform an additional exploratory analysis to explore the predictive performance of different features and establish whether lost parcels have distinctive characteristics from normal parcels. For all numerical features, the boxplots and histograms are compared for two datasets, one containing lost products (LI) and the other exclusively normal products (not LI). In Fig. \ref{fig: Pareto}, the boxplot of \textit{size\_length}, i.e. length of the product (cm) is shown for both groups. It can be observed that there are some differences in the distribution of this predictive feature. It is observed that lost products are, on average smaller than normal products. Similar differences are observed for other numerical features, which suggests that these features have some predictive power, despite no linear correlations were observed. The same analysis is performed for the boolean features. Both groups' mean and standard deviation slightly differ, indicating that the boolean features can be used for later predictive objectives. An example of this difference is visualized in Fig. \autoref{fig: Pareto Close up}. The figure implicitly shows that more parcel loss occurs at B2B customers than regular customers.


The same exploratory analysis is performed for categorical features. 
We observe that the percentage of products that belongs to the team ‘Telefoon, Tablets \& Accessoires', which are phones, tablets, and accessories, is significantly higher (42,73\%) for LI's, compared to the percentage in the group of normal (not LI) products (24,20\%). It seems thus that relatively many phones, tablets, and accessories are lost. Similar observations are made for other categorical features. For instance, there is a difference between the relative amount of parcels lost in area Wallonia and Flanders. The features, province and region, seem to be important predictors of LI's. In addition, the payment method appears to be an important predictor of LI's. On the other hand, gender, type of order, and type of outlet seem to have less predictive power because no differences are observed.  

Several sequential data preprocessing steps are performed on the collected raw dataset, including data cleaning and transformation. Data cleaning involves handling missing values and reducing data noise by handling data errors and outliers. The data transformation includes the encoding of features to the correct data format and data generation. The output of the data preparation phase is a dataset that can be used in the next phase, modeling. 

We generate an additional feature named DateDeliveryISOMonth, representing the month of delivery, as it was observed that the month seemed to impact parcel loss quantity. 
Moreover, it was observed that orders placed on Saturday resulted more often in parcel loss. Hence, the feature weekday is generated, representing the day of the week on which the order is placed. 

As many machine learning models require numerical values as inputs, we use one-hot encoding to transform all categorical features to numerical ones. 
This, however, results in high-dimensional datasets, especially if many different feature values are observed. Due to the large dataset size, it is chosen to relabel the categorical features into a maximum of 20 categories. This data transformation results in a loss of information but decreases the probability of overfitting. 

After data preprocessing, the dataset consists of 854.898 rows with 226 predictive features and one target variable. Each row represents one parcel, that can contain one or multiple products. This dataset used for modeling contains 2174 lost parcels, which represents 0,25\% of the total dataset. 
In \autoref{tab: dataset for modeling} a snapshot of the dataset after preprocessing is presented to provide an indication of the dataset that is used for modeling. The table shows the first parcel (index 0) was lost (Is\_lost\_item = 1), while the next two parcels were not (Is\_lost\_item = 0). 

\begin{table}[H]
\centering
\scriptsize
\begin{tabular}{|c|c|c|c|c|c|c|c|c|c|}
\hline
\textbf{} & \textbf{\begin{tabular}[c]{@{}c@{}}size\_\\ width\end{tabular}} & \textbf{\begin{tabular}[c]{@{}c@{}}stock\_\\ value\end{tabular}} & \textbf{quantity} & \textbf{....} &  \textbf{\begin{tabular}[c]{@{}c@{}}team\_name\_\\ Team Audio\end{tabular}} & \textbf{\begin{tabular}[c]{@{}c@{}}brand\_name\\ \_Acer\end{tabular}} & \textbf{\begin{tabular}[c]{@{}c@{}}has\_\\ phone\end{tabular}} & \textbf{\begin{tabular}[c]{@{}c@{}}Is\_lost\\ \_item\end{tabular}} \\ \hline
0         & 2.230                                                           & 6,561                                                            & 1                 & ....                                                                                 & 0                                                                          & 1                                                                     & 1                                                              & 1                                                                  \\ \hline
1         & 1.705                                                           & 4,667                                                            & 1                 & ....                                                                                & 0                                                                          & 0                                                                     & 1                                                              & 0                                                                  \\ \hline
2         & 2.501                                                           & 4,976                                                            & 1                 & ....                                                                                 & 0                                                                          & 0                                                                     & 1                                                              & 0                                                                  \\ \hline
\end{tabular}
\caption{Example of dataset after preprocessing. }
\label{tab: dataset for modeling}
\end{table}

Finally, it should be noted that the continuous features are log-transformed to improve the predictive performance in the modeling phase. Nevertheless, these feature values should be transformed back during business evaluation to draw relevant conclusions.

\section{Prediction Methods}
\label{sec:methods}
In this section, we propose two ML approaches for predicting parcel loss in last-mile delivery with imbalanced dataset. First, we study commonly used data imbalance techniques applied to standard supervised techniques. We name this method Data Balance Techniques with Supervised Learning (DBSL). Then, we introduce our novel Deep Hybrid Ensemble Learning (DHEL) method which utilizes an autoencoder and ensemble classifiers to more effectively handle imbalanced data and improve predictive performance. 

A brief introduction to the standard supervised learning algorithms and autoencoders used in our study, as well as their variants, is provided in the Appendix. For further background information on these techniques, we refer the reader to a seminal book by \cite{goodfellow2016deep}. Details of DBSL and DHEL are provided in subsequent subsections.


\subsection{Data Balance Techniques with Supervised Learning (DBSL)} 
We study the performance of various data imbalance correction techniques combined with standard supervised learning algorithms for the parcel loss prediction task. Specifically, we experiment with four different resampling methods: Random Undersampling (RU), Near-Miss Undersampling, Random Undersampling Boosting (RUSBoost), and UnderBagging (UB). Each technique is used to balance the class distribution of the imbalanced parcel loss dataset. 


\textbf{Random Undersampling (RU)} is the most naïve method of undersampling, which tries to balance the class distributions by randomly selecting fewer examples from the majority class~\citep{kotsiantis2003mixture}. RU is an efficient method; however, it can lead to the random removal of informative samples from the data.

\textbf{Near-Miss Undersampling} undersamples majority class instances based on the distance of the majority class to minority
instances~\citep{mani2003knn}. There are three versions of NM. In this paper, we uses NearMiss-1 because the other two versions are computationally expensive. NearMiss-1 (hereafter as NM) selects samples from the majority class close to three of the closest examples from the minority class and removes them. NM not only determines the most representative instances from the majority class but also covers the
most easily misclassified ones~\citep{bao2016boosted}.

\textbf {Random Undersampling Boosting (RUSBoost) } is a technique that combines random undersampling with boosting. Random undersampling balances class distributions by randomly removing majority class samples~\citep{seiffert2009rusboost}. Boosting then trains sequential weak learners that focus on misclassified samples from previous models. The combined votes from all weak learners produce an ensemble model. Random undersampling reduces overfitting to the majority class, while boosting combines weak learners to produce a strong final model.

\textbf{UnderBagging (UB)} incorporates the strength of RU with bagging.  It first balances class distributions through undersampling the majority class~\citep{raghuwanshi2018underbagging}.  It then trains multiple models on random subsets of the balanced data, as in bagging. The goal is to benefit from both undersampling to balance classes and bagging to reduce variance through combining diverse models.

\begin{figure}[H]
    \centering
    \includegraphics[width = 0.85\textwidth]{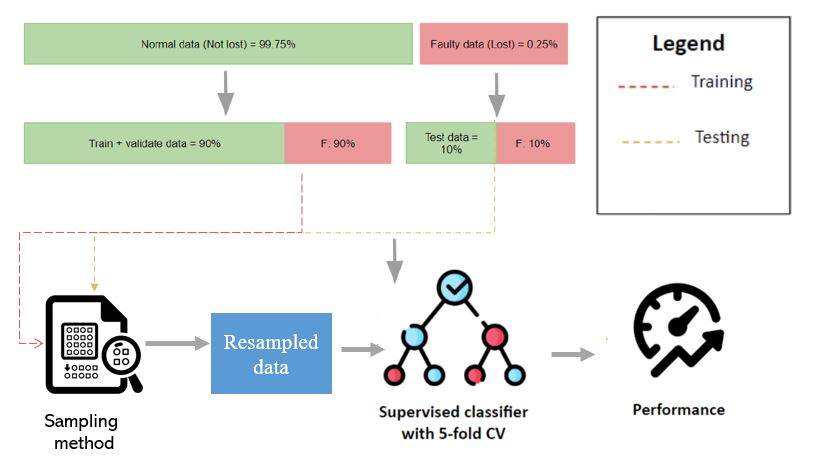}
    \caption{An overview of Data Balance Techniques with Supervised Learning(DBSL)}
    \label{fig:dbsl}
\end{figure}

The balanced samples generated via these resampling strategies are then used to train five different supervised learning models: Decision Tree (DT), Random Forest (RF), Extreme Gradient Boosting (XGB), Logistic Regression (LR), and Support Vector Machine (SVM). These represent both linear and non-linear classification algorithms that are commonly applied. Figure~\ref{fig:dbsl} provides an overview of DBSL training and testing procedure. 

\subsection{Deep Hybrid Ensemble Learning (DHEL)}
Autoencoders (AE) can learn representations of data by attempting to reconstruct their inputs. The reconstruction error measures how well each sample was represented in the latent space. Samples with higher errors may indicate anomalies or minorities that deviate from the majority. In the proposed Deep Hybrid Ensemble Learning (DHEL) approach, we employ an AE to extract features from the imbalanced parcel loss dataset consisting of only normal data (i.e. not lost). Rather than using the encoded features directly, we leverage the reconstruction error vector output by the decoder layer. These error measurements are then provided as inputs to an ensemble of supervised learning models. By utilizing the reconstruction failures highlighted by the AE, the classifiers aim to better identify hard to represent minority parcel losses. We examine DHEL using different autoencoder architectures, including variational and denoising variants, to determine the most effective configuration. A brief explanation of each autoencoder type is provided in the Appendix. The developed hybrid approach proposed in this research is inspired from~\cite{lin2021credit}, which uses it for credit card fraud detection and study on random forests with AE.

\begin{figure}[H]
    \centering
    \includegraphics[width = 0.9\textwidth]{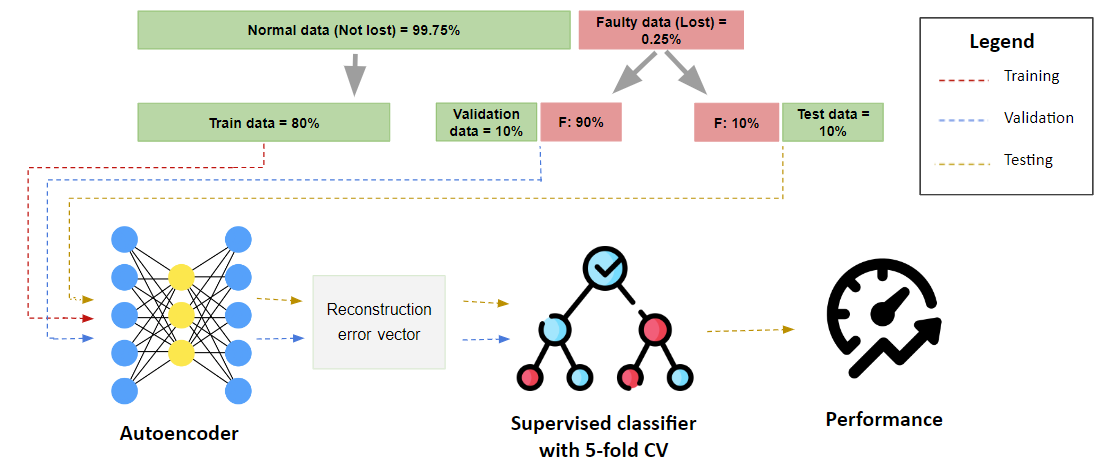}
    \caption{An overview of Deep Hybrid Ensemble Learning (DHEL) with reconstruction error}
    \label{fig: deep hybrid}
\end{figure}

\autoref{fig: deep hybrid} illustrates that in the deep hybrid approach with reconstruction error vector the training set contains only normal data instances and is solely used to train the AE network. The autoencoder reproduces the validation set as reconstruction error vectors \textit{$E_v$}, seen as the supervised classifier's input and ‘training' data. The actual training data \textit{$D_{train}$} is only used to train the AE to prevent data leakage. Hence, the validation data \textit{$D_{validate}$} is used for training and hyperparameter optimization of the supervised classifier \textit{c}. Random search is combined with repeated stratified 5-fold cross-validation to retrieve the optimal hyperparameters for each supervised classifier \textit{c}. The stratified 5-fold cross-validation ensures that all five folds contain sufficient anomalous data instances. Stratification reduces the estimate's variance, making each fold representative of the whole dataset. Once the final hyperparameters are found, the complete validation set \textit{$D_{validate}$} is afterwards used to train the classifier \textit{c} as this is expected to improve the model's generalization ability. The test set \textit{$D_{test}$} is only used to evaluate the final performance of the deep hybrid model.

\section{Performance of prediction models}
\label{sec:performance}
The developed models are evaluated on the retrieved and preprocessed dataset described in Section \ref{sec:data}. 
The autoencoders are implemented using Keras Tensorflow \citep{chollet2018keras}. The supervised classification models are implemented using the sci-kit learn library~\citep{pedregosa2011scikit}. The training and evaluation of the models is performed on a Processor Intel(R) Core i7-4710MQ. 

We use the confusion matrix, shown in Table \ref{tab:confusion}, to visualize the performances of different prediction models. As we have a binary classification problem, the outcomes of the prediction models are either positive (Class 1) or negative (Class 0). In our case, a positive prediction indicates the parcel being lost. We use the following evaluation measures to assess and compare prediction models. \emph{Precision}, defined by $\frac{TP}{TP+FP}$ (see Table \ref{tab:confusion}), measures the relative amount of parcels predicted as lost that is actually lost.  \emph{Recall}, computed as $\frac{TP}{TP+FN}$, measures the relative amount of lost parcels that are detected. 
Moreover, we adopt the \emph{Balanced Accuracy} (BA) metric to evaluate the performance of prediction models across both classes in the presence of imbalanced data. BA is computed as $\frac{1}{2}(\frac{TP}{P}+\frac{TN}{N})$.  
In addition, the \emph{Receiver Operating Characteristics (ROC) curve} is used to obtain the Area Under Curve score (AUC). The ROC-AUC summarizes the True Positive Rate (TPR) and False Negative Rate (FNR) of all different AUC’s from both classes’ ROC. It can take a value between 0 and 1, where an AUC of 0,5 suggests the ROC falls on the diagonal and the prediction has no discriminatory ability to classify. An AUC of 1,0 means that the classifier is perfect at discrimination. Finally, we also show the True Negative Rate (TNR) value to measure how the model accurately detects the normal (not lost) parcels. 
These metrics align well with our use case, characterized by highly imbalanced data. We use the Balanced Accuracy (BA) and AUC (ROC\_AUC) to choose the best models as they provide a more comprehensive representation of the model's performance. Other measures serve as additional explanations of the model's performance. 
\begin{table}[h]
    \centering
    \begin{tabular}{l|l|c|c|}
        \multicolumn{2}{c}{} & \multicolumn{2}{c}{Predicted class} \\
        \cline{3-4}
        \multicolumn{2}{c|}{} & Positive (Class 1)& Negative (Class 0) \\ 
        \cline{2-4}
        \multirow{2}{*}{Actual class} & Positive & True Positive $TP$ & False Negative $FN$ \\
        \cline{2-4} 
        & Negative & False Positive $FP$ & True Negative $TN$ \\
        \cline{2-4}
    \end{tabular}
    \caption{Confusion matrix to illustrate prediction performance.}
    \label{tab:confusion}
\end{table}

\subsection{Performance of the Business Rules}
We use the current business insurance rules at Company X as the baseline method to compare to various machine learning methods. 
The predictive performance of the business rules model is established using the 10\% test set containing 85,273 normal parcels and 217 lost parcels. \autoref{tab: baseline} presents the predictive performance of the business rules model and \autoref{fig: confusion matrix business rules} shows the corresponding confusion matrix.  In total, 42,4\% of the lost parcels are correctly detected. The balanced accuracy (BA) and ROC-AUC score show identical values 0,562 because the defined decision rules cannot predict probabilities but only binary outcomes. Therefore, no different thresholds can be obtained, making the ROC-AUC score identical to the balanced accuracy score. The ROC-AUC score is just above 0,5 indicating that the performance is slightly better than random guessing.


\begin{table}[H]
\centering
\begin{tabular}{lcccccc}
\toprule
\textbf{Model}  & \textbf{TNR} & \textbf{Precision} & \textbf{Recall} & \textbf{ROC-AUC} & \textbf{BA} \\ \midrule
BR & 0,699 & 0,004     & 0,424  & 0,562   & 0,562        \\ \bottomrule
\end{tabular}
\caption{Business rules classification results on the 10\% test set.}
\label{tab: baseline}
\end{table}


\begin{table}[h]
    \centering
    \begin{tabular}{|l|c|c|}\hline
  & Positive (Class 1)& Negative (Class 0) \\ \hline
   Positive & 92 & 125 \\ \hline
          Negative & 25567 & 59706 \\ \hline
  \end{tabular}
    \caption{Confusion matrix of business rules model classification model on the 10\% test set.}
    \label{fig: confusion matrix business rules}
\end{table}

\subsection{Data Balance Techniques with Supervised Learning (DBSL) Results}
We developed multiple supervised models and evaluated their predictive performance. Different strategies such as sampling, and ensemble learning were evaluated as improvements to traditional classifiers to address the class imbalance. 
The sampling strategy is defined as the distribution of majority samples against minority samples. \textit{Random Undersampling (RU)} and\textit{ Near-Miss Undersampling (NM)} are used as sampling methods whereas \textit{RUSBoost} and \textit{UnderBagging (UB)} are examined as ensemble methods, which incorporate random undersampling in the algorithm. The data is scaled for the LR and SVM models, as scaling is unnecessary for tree-based methods. 
Initially, stratified 5-fold cross-validation combined with sk-learn randomized search is performed on the training set to perform hyperparameter tuning for all models. The best hyperparameters are selected based on the balanced accuracy score. These optimized models are trained again on the entire 90\% training set and evaluated on the 10\% test set.

  \begin{figure}[!ht]
    \subfloat[RU\label{subfig:supervised_ru_roc}]{%
      \includegraphics[width=0.25\textwidth]{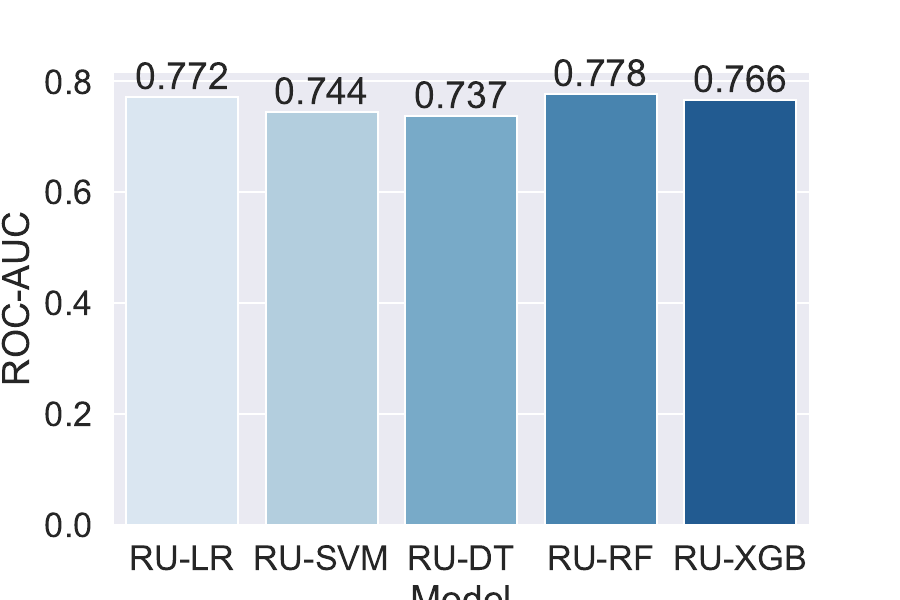}
    }
        \subfloat[UB\label{subfig-2:supervised_ub_roc}]{%
      \includegraphics[width=0.25\textwidth]{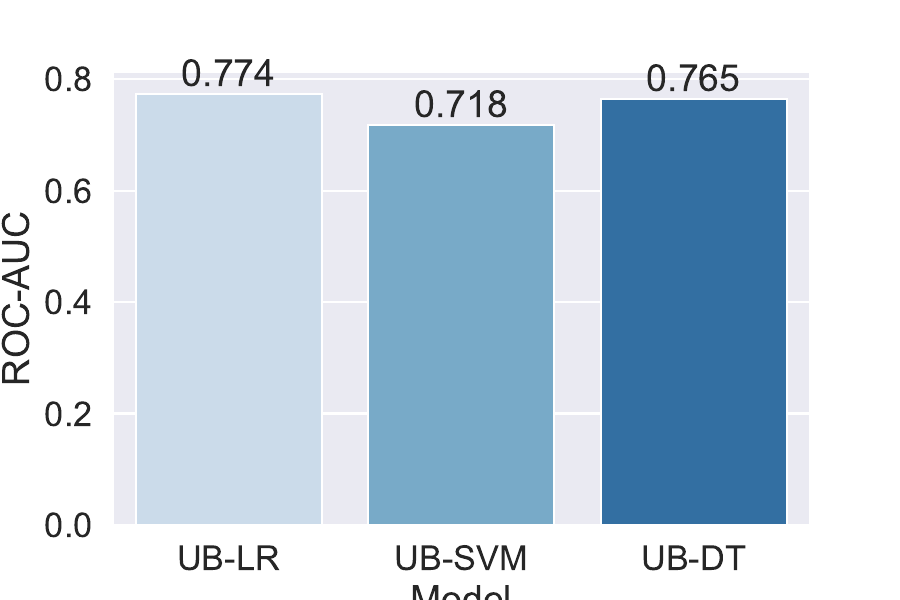}
    }
        \subfloat[NM\label{subfig-2:supervised_nm_roc}]{%
      \includegraphics[width=0.25\textwidth]{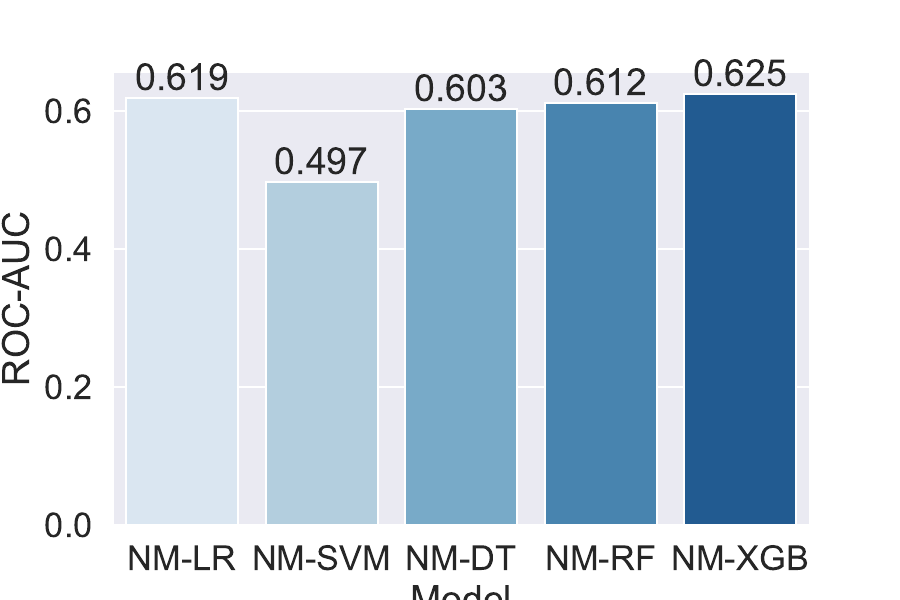}
    } 
    \subfloat[RUSBoost\label{subfig-2:supervised_boost_roc}]{%
      \includegraphics[width=0.25\textwidth]{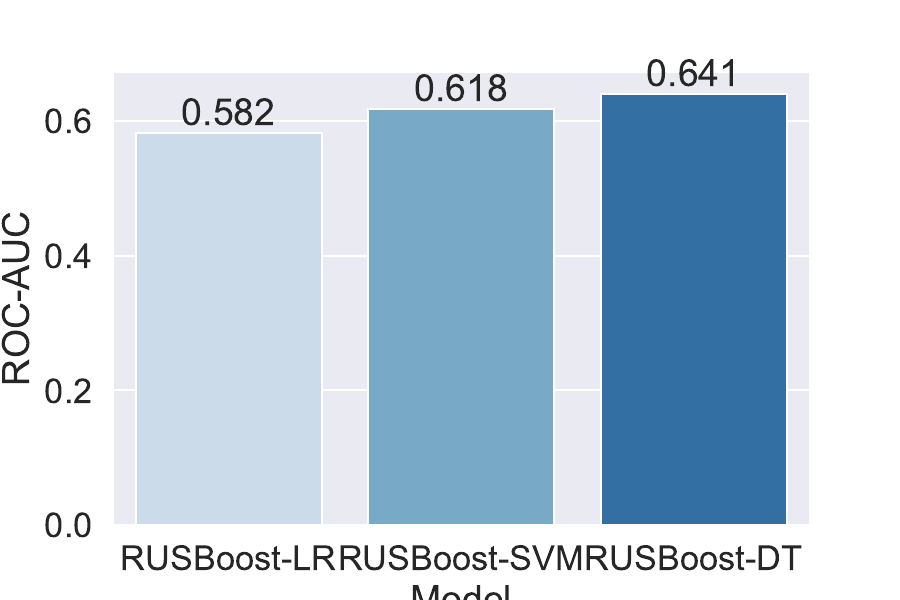}
    }
    \caption{\textbf{ROC-AUC scores of optimised models on 10\% test set}. Random sampling (RU), Underbagging (UB). Near miss (NM) undersampling, Random Undersampling Boosting (RUSBoost)}
    \label{fig:supervised-roc-auc}
  \end{figure}

  \begin{figure}[!ht]
    \subfloat[RU\label{subfig:supervised_ru_bac}]{%
      \includegraphics[width=0.25\textwidth]{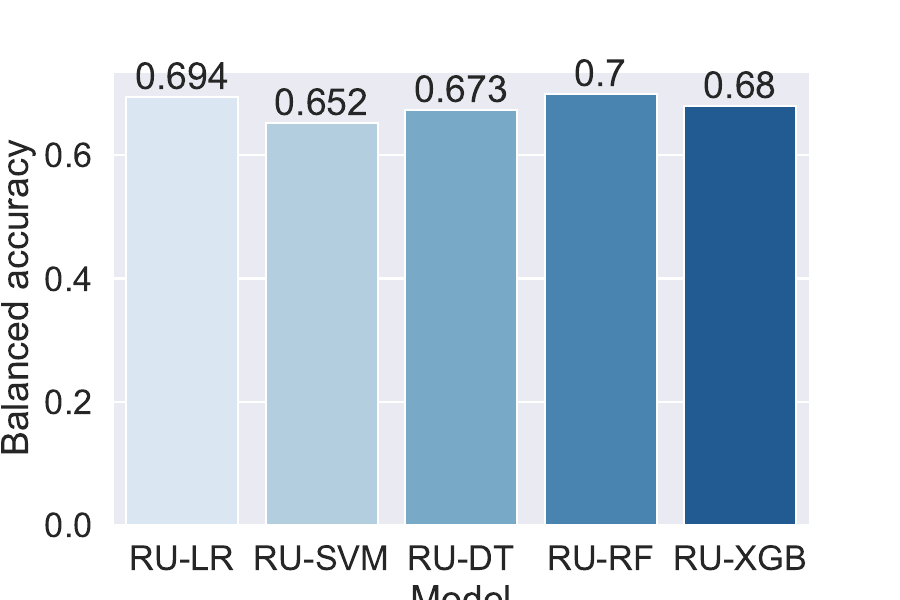}
    }
    \subfloat[UB\label{subfig-2:supervised_ub_bac}]{%
      \includegraphics[width=0.25\textwidth]{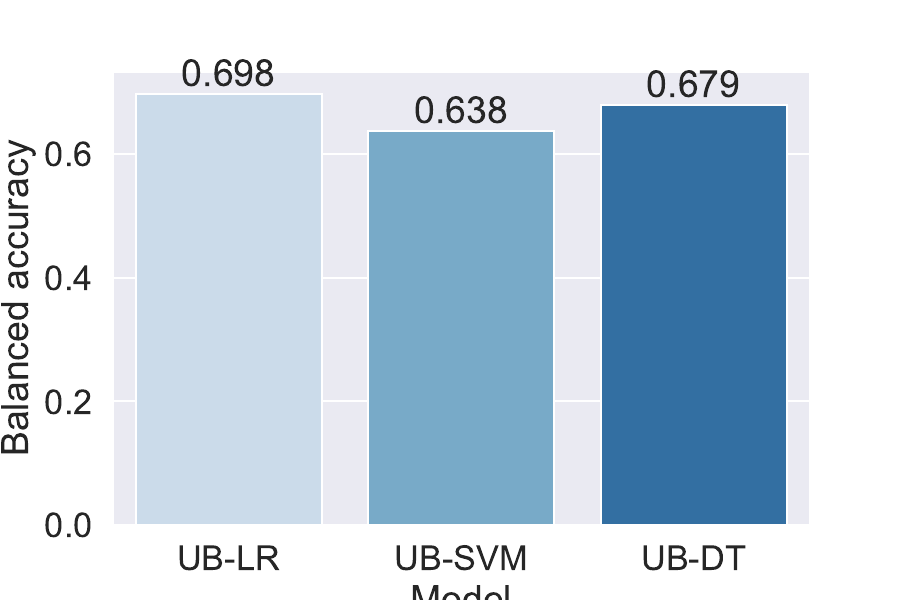}
    } 
        \subfloat[NM\label{subfig-2:supervised_nm_bac}]{%
      \includegraphics[width=0.25\textwidth]{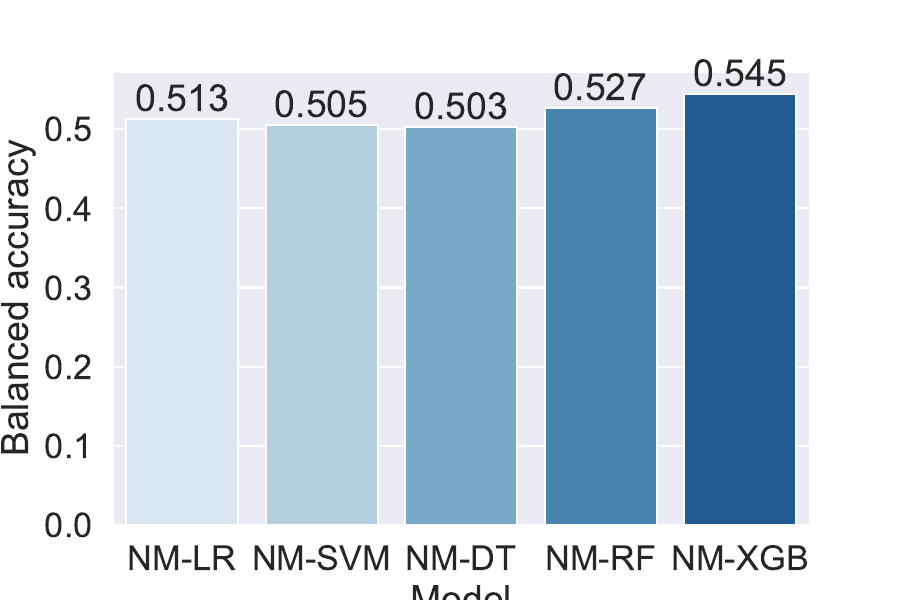}
    }
    \subfloat[RUSBoost\label{subfig-2:supervised_boost_bac}]{%
      \includegraphics[width=0.25\textwidth]{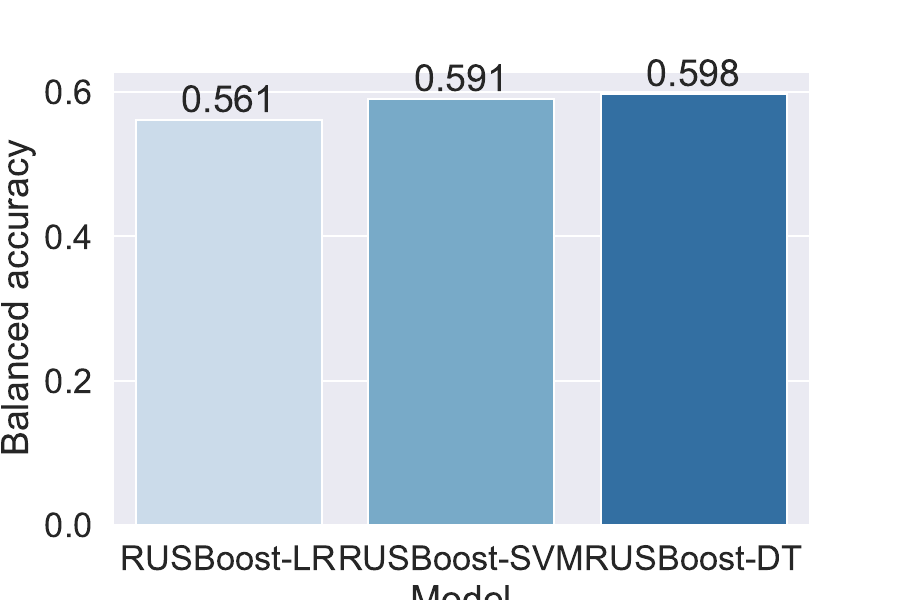}
    }
    \caption{\textbf{Balanced accuracy scores of optimised models on 10\% test set}. Random sampling (RU), Underbagging (UB). Near miss (NM) undersampling, Random Undersampling Boosting (RUSBoost)}
    \label{fig:supervised-bac}
  \end{figure}



Figure~\ref{fig:supervised-roc-auc} and~\ref{fig:supervised-bac} reports the ROC-AUC and balanced accuracy scores of tree-based models, along with LR and SVM. Note that we did not combine underbagging (UB) and RUSBoost with random forest and XGBoost as they are ensemble methods of the decision tree itself. It can be observed that the LR model shows good predictive performance while being a relatively simpler model. The DT model consistently performs better than the business rules, which indicates that machine learning improves decision-making on parcel loss. With random sampling (RU), all the models, except for the business rule, show comparable performance with RF model securing 0,778 ROC-AUC and 0,7 balanced accuracy score. The UB-LR model performs best, detecting 65,9\% of lost parcels with  ROC-AUC equals 0,774 and balanced accuracy of 0,698. The undersampling approaches, both NM-undersampling and RUS-Boost, do not improve the predictive performance of any model.  NM is likely overfitting, whereas RUSBoost focuses on the wrongly misclassified normal parcels, instead of the wrongly misclassified lost parcels. The majority voting systems in UnderBagging seems to prevent the classifiers from overfitting, causing an increase in performance in the ensemble model over the standard classifiers.

\autoref{tab: supervised results} shows the predictive performance of optimized models. Remarkably, all different machine learning techniques offer better performance in combination with RU.  In general, it can be observed that parcel loss prediction is a complex problem, as the precision score is for all models below 0,02, indicating that out of all predicted lost parcels, less than $2$\% is lost. In general, missing a TP is more costly than missing a TN. Therefore, a relatively low precision seemed less critical. The UB-SVM model shows the worst performance, resulting in a relatively low ROC-AUC 0,744, and balanced accuracy score 0,652. However, the precision is slightly higher than the other classifiers 0,021.

\begin{table}[H]
\centering
\begin{tabular}{lrcccc}
\toprule
\textbf{Model} & \multicolumn{1}{c}{\textbf{TNR}} & \textbf{Precision} & \textbf{Recall} & \textbf{ROC-AUC} & \textbf{BA} \\ \midrule
BR & 0,700                             & 0,004              & 0,424           & 0,562            & 0,562                      \\ 
UB-DT          & 0,888                             & 0,011              & 0,470           & 0,765            & 0,679                      \\ 
RU-RF          & 0,912                             & 0,014              & 0,488           & 0,778            & 0,700                      \\ 
RU-XGB         & 0,930                             & 0,015              & 0,433           & 0,766            & 0,681                      \\ 
UB-LR          & 0,743                             & 0,006              & 0,659           & 0,774            & 0,698                      \\ 
RU-SVM         & 0,958        & 0,021              & 0,346           & 0,744            & 0,652                      \\ \bottomrule
\end{tabular}

\caption{Performance of optimized models on 10\% test set.}
\label{tab: supervised results}
\end{table}

\subsection{Deep Hybrid Ensemble Learning (DHEL) results}
We analyzed the performance of our proposed Deep Hybrid Ensemble Learning (DHEL) approach for predicting parcel losses. A key aspect of an effective AE model is appropriately selecting hyperparameters and thresholds. Therefore, we first detail our process for hyperparameter tunning and threshold setting followed by a performance evaluation of different models. 
\subsubsection{Hyperparameter tunning and threshold setting of AEs}
We used a validation set to optimize the hyperparameters of AEs. The validation set contains both normal and anomalous instances. The optimal architecture and hyperparameters per autoencoder are determined using a random search. \autoref{tab: optimal hyperpar autoencoders} provides an overview of optimal parameter configuration of AE, VAE and DAE.

\begin{table}[H]
\centering
\begin{tabular}{lccc}
\toprule
 & \textbf{AE}      & \textbf{VAE}     & \textbf{DAE}       \\ \midrule
Optimizer                              & Adam    & Adam    & Adam      \\ 
Loss function                          & MSE     & MSE     & MSE       \\ 
Activation function input/output layer & ReLU    & Sigmoid & ReLU      \\
Activation function hidden layers      & Sigmoid & eLU     & Sigmoid   \\ 
Nodes per hidden layer                 & 8-4-8   & 8-4-8   & 8-4-2-4-8 \\ 
Dropout rate                           & 0.2     & 0.5     & 0.1       \\ 
Amount of input features               & 25      & 15      & 25        \\ 
Latent dimensions                      & -       & 4       & -         \\ 
Standard deviation input noise         & -       & -       & 0,2       \\ \bottomrule
\end{tabular}
\caption{Optimal network architecture and hyperparameters for three types of autoencoders, found during parameter optimization on the 10\% validation set}
\label{tab: optimal hyperpar autoencoders}
\end{table}

To classify instances as anomalous, a threshold value is required. The threshold value is obtained using the MSE of the reconstruction error from the validation set.  As a rule of thumb, the score of 3,5 is used as a cut-off value, meaning all instances having a modified Z-score above 3,5 are classified as anomalies. However, manually adjusting the threshold per model can overestimate the models' performance. Finding a suitable threshold value requires a tradeoff between precision and recall. Improving recall is obtained by lowering the threshold while recovering precision requires the opposite. 

The balanced accuracy metric inherits the tradeoff between precision and recall. Therefore, different threshold values can be evaluated on the balanced accuracy score obtained on the validation set and thus can be used to determine the optimal threshold value. \autoref{tab: reconstruction threshold } presents the predictive performance of the regular AE on the validation set for different threshold values. It can be observed that higher threshold values result in lower recall but higher precision scores and vice-versa. Furthermore, is is observed that the ROC-AUC score is equal for all different threshold values, which was expected because the ROC-AUC score is calculated by the area under the precision recall curve. 


\begin{table}[H]
\centering
\begin{tabular}{crcccc}
\toprule
\textbf{Threshold} & \multicolumn{1}{c}{\textbf{TNR}} & \textbf{Precision} & \textbf{Recall} & \textbf{ROC-AUC} & \textbf{BA} \\ \midrule
0,0200             & 0,000                             & 0,022              & 1,000           & 0,611            & 0,500                      \\ 
0,0300             & 0,012                             & 0,023              & 0,992           & 0,611            & 0,502                      \\ 
0,0350             & 0,397                             & 0,026              & 0,709           & 0,611            & 0,554                      \\ 
0,0380             & 0,697                             & 0,034              & 0,467           & 0,611            & 0,582                      \\ 
\textbf{0,0391}             & \textbf{0,777}                             & \textbf{0,040 }             & \textbf{0,406}           & \textbf{0,611}            & \textbf{0,592}                      \\ 
0,0420             & \multicolumn{1}{c}{0,925}        & 0,068              & 0,239           & 0,611            & 0,582                      \\ 
0,0500             & \multicolumn{1}{c}{0,995}        & 0,226              & 0,061           & 0,611            & 0,528                      \\ \bottomrule
\end{tabular}
\caption{The 10\% validation set performance of the regular autoencoder for different threshold values} 
\label{tab: reconstruction threshold }
\end{table}

For the regular AE, the threshold value of 0,0391 is set based on the validation set outcomes, resulting in a precision of 0,040, recall of 0,406, ROC-AUC score of 0,611, and balanced accuracy of 0,592. Following the same approach, the threshold value of 0,0015 and 0,0350 is set for VAE and DAE, respectively. 

\subsubsection{Results}
\autoref{tab: Reconstruction error deep hybrid results} provide the performance of deep hybrid models with optimized hyperparameters on the unseen 10\% test set. The deep hybrid models using the logistic regression (LR) and SVM classifier show relatively poor predictive performance. The SVM algorithm was already observed to be underperforming during supervised classification, 
and the poor performance of LR can be due to the interdependence of the reconstruction errors and an absence of multi-collinearity. The reconstruction error vector obtained from the autoencoder causes the input data to be correlated. Therefore, it is expected that the relatively low performance of the deep hybrid LR model is due to the transformation caused by AE models. The XGBoost and RF classifiers seem to outperform the DT model. Boosting shows slightly lower performance compared to bagging, which was also observed during supervised classification. The best performance is obtained with a regular autoencoder and random forest (AE-RF). The AE-RF model shows promising results, with an ROC-AUC score equals to $0,759$, a balanced accuracy score of 0,704, and in total, correct detection of 51,3\% of all lost parcels. Moreover, it can be observed that all deep hybrid models significantly outperform the business rules. 

The more robust VAE and DAE do not improve the predictive performance, as all deep hybrid models show lower predictive performance than the regular AE ensembles. It is expected that the added random noise in the reconstruction error of the DAE and  VAE  causes the classifier to focus on the wrong features during training. Moreover, the VAE was intended to make the deep hybrid model more robust by sampling from the Gaussian distribution of the latent representations. However, it is expected that similar to the DAE this additional noise in the input data causes the model to focus more on random noise instead of the noise in the minority class instances. 

\begin{table}[t!]
\centering
\small
\begin{tabular}{ccccccc}
\toprule
\multicolumn{1}{c}{\textbf{Autoencoder}} & \textbf{Classifier} & \textbf{TNR} & \textbf{Precision} & \textbf{Recall} & \textbf{\begin{tabular}[c]{@{}c@{}}ROC\_\\ AUC\end{tabular}} & \textbf{\begin{tabular}[c]{@{}c@{}}Balanced\\ Accuracy\end{tabular}} \\ \midrule
\multicolumn{1}{c}{AE}                   & RF                  & 0,878        & 0,010              & 0,513           & \textbf{0,759 }                                                       & \textbf{0,704 }                                                               \\ 
\multicolumn{1}{c}{AE}                   & XGB                 & 0,984        & 0,051              & 0,330           & 0,757                                                        & 0,657                                                                \\ 
\multicolumn{1}{c}{AE}                   & LR                  & 0,736        & 0,005              & 0,505           & 0,669                                                        & 0,620                                                                \\ 
\multicolumn{1}{c}{AE}                   & SVM                 & 0,877        & 0,007              & 0,304           & 0,644                                                        & 0,592                                                                \\ 
\multicolumn{1}{c}{AE}                   & DT                  & 0,860        & 0,009              & 0,486           & 0,733                                                        & 0,673                                                                \\ 
\multicolumn{1}{c}{VAE}                  & RF                  & 0,093        & 0,015              & 0,395           & 0,732                                                        & 0,665                                                                \\ 
\multicolumn{1}{c}{VAE}                  & XGB                 & 0,984        & 0,048              & 0,321           & 0,732                                                        & 0,652                                                                \\ 
\multicolumn{1}{c}{VAE}                  & LR                  & 0,784        & 0,006              & 0,482           & 0,671                                                        & 0,633                                                                \\ 
\multicolumn{1}{c}{VAE}                  & SVM                 & 0,993        & 0,012              & 0,032           & 0,631                                                        & 0,512                                                                \\ 
\multicolumn{1}{c}{VAE}                  & DT                  & 0,952        & 0,019              & 0,358           & 0,684                                                        & 0,655                                                                \\ 
\multicolumn{1}{c}{DAE}                  & RF                  & 0,913        & 0,012              & 0,432           & 0,728                                                        & 0,672                                                                \\ 
\multicolumn{1}{c}{DAE}                  & XGB                 & 0,985        & 0,048              & 0,298           & 0,725                                                        & 0,642                                                                \\ 
\multicolumn{1}{c}{DAE}                  & LR                  & 0,799        & 0,005              & 0,417           & 0,650                                                        & 0,608                                                                \\ 
\multicolumn{1}{c}{DAE}                  & SVM                 & 0,994        & 0,012              & 0,198           & 0,608                                                        & 0,580                                                                \\ 
\multicolumn{1}{c}{DAE}                  & DT                  & 0,922        & 0,013              & 0,399           & 0,704                                                        & 0,660                                                                \\ 
\multicolumn{2}{c}{Business rules}                             & 0,700        & 0,004              & 0,424           & 0,562                                                        & 0,562                                                                \\ \bottomrule
\end{tabular}
\caption{10\% Test set results (85490 instances) of all deep hybrid models using the reconstruction error as input}
\label{tab: Reconstruction error deep hybrid results}
\end{table}

\subsection{Comparison of models from DBSL and DHEL}
\autoref{tab: final result comparison} shows the average performance on a 10\% test set of the best models of different learning approaches. It can be noted that all models improve classification performance over the current business rules, which validates the choice for the company  to use machine learning in parcel loss prediction. The semi-supervised VAE shows the lowest classification performance. It was concluded that a rigid threshold value on the MSE of the reconstruction does not provide enough flexibility to perform accurate classification with autoencoders. 

Random forest is observed to be the most optimal machine learning algorithm for parcel loss prediction as both the supervised and deep hybrid model uses the algorithm to perform classification. The deep hybrid AE-RF model uses the reconstruction error vector as input to the random forest model to perform classification. It can be observed that the predictive performance is similar for the RU-RF and AE-RF models. The supervised RU-RF model shows the highest average ROC-AUC performance (0,768). However, the average recall and balanced accuracy are higher for the deep hybrid AE-RF model. Hence, the results shows that the most optimal model for parcel loss prediction is the deep hybrid AE-RF model. This model detects on average $55,4\%$ of all lost parcels, the average precision equals 0,010, average ROC-AUC equals 0,763 and the average balanced accuracy equals 0,701. Table \ref{tab:confusion matrix RURF} and Table \ref{tab:confusion matrix RURF} show the confusion matrix obtained from the 10\% test set for both RU-RF and deep hybrid AE-RF models. It can be observed that the differences in actual predictions are minimal. The AE-RF model detects more true negatives, which comes at the cost of slightly more false negatives. 

\begin{table}[tb]
\centering
\begin{tabular}{ccccc}
\toprule
\multicolumn{1}{l}{Model} & \multicolumn{1}{c}{\textbf{Precision}}                          & \multicolumn{1}{c}{\textbf{Recall}}                    & \multicolumn{1}{c}{\textbf{ROC-AUC}}                            & \multicolumn{1}{c}{\textbf{BA}} \\ \midrule

Business rules         & \multicolumn{1}{c}{0,004 $\pm$ 0,000}   & \multicolumn{1}{c}{0,422 $\pm$ 0,018} & \multicolumn{1}{c}{0,561 $\pm$ 0,009}  & \multicolumn{1}{c}{0,561 $\pm$ 0,009}   \\ 

RU-RF                  & \multicolumn{1}{r}{0,010 $\pm$ 0,001}  & \multicolumn{1}{r}{0,478 $\pm$ 0,026}                        & \multicolumn{1}{r}{0,768 $\pm$ 0,036}                       & \multicolumn{1}{c}{0,695 $\pm$ 0,035}       \\ 


AE-RF                  & \multicolumn{1}{r}{0,010 $\pm$  0,000 }                       & \multicolumn{1}{r}{0,554 $\pm$ 0,039}                        & \multicolumn{1}{r}{0,763 $\pm$ 0,031}                        & \multicolumn{1}{c}{0,701 $\pm$ 0,027 }      \\ \bottomrule
\end{tabular}
\caption{Evaluation of the average performance of the best-performing models per learning approach on ten different 10\% test sets with 85,490 instances}
\label{tab: final result comparison}
\end{table}

\begin{table}[ht]
\centering
    \begin{tabular}{|l|c|c|}\hline
  & Positive (Class 1)& Negative (Class 0) \\ \hline
   Positive & 106 & 111 \\ \hline
          Negative & 7476 & 77797 \\ \hline
  \end{tabular}
    \caption{Confusion matrix of RU-RF model on the test set.}
    \label{tab:confusion matrix RURF}
\end{table}



\begin{table}[h]
    \centering
     \begin{tabular}{|l|c|c|}\hline
  & Positive (Class 1)& Negative (Class 0) \\ \hline
   Positive & 112 & 105 \\ \hline
          Negative & 10443 & 74830 \\ \hline
  \end{tabular}
    \caption{Confusion matrix of AE-RF model on the test set.}
    \label{tab:confusion matrix AERF}
\end{table}

\section{Business Value and Managerial Implications}
\label{sec:business}
We discuss the business value and insights from the developed predictive models in the section. 
\subsection{Insurance decision-making with predictive models}
The developed predictive model is intended to be used at Company X as an alternative to the current insurance rules. That is, for each parcel to be delivered, the model predicts whether it will be lost during last-mile delivery. If the model predicts that the parcel is going to be lost, the insurance is used and vice-versa. 

In this way, we can evaluate the monetary impact of the predictive models using misclassification costs.  In addition, the company would like to increase understanding of parcel loss in the last-mile delivery and explore how models can be used to prevent parcel loss. Hence, we define the second business evaluation metric as the interpretability of the model, which relates to the business information that can be extracted. we use the methods from Explainable AI (XAI) for this purpose. 


\subsection{Misclassification costs}
The misclassification costs are computed based on the number of incorrectly classified parcels: false positives (FP) and false negatives (FN).  
FP are predictions of parcels that are predicted to become lost, while in reality, these parcels were not lost. Translated to the business process, these parcels would have been unnecessarily insured. The  corresponding insurance cost function \textit{IC($S_i$,$P_i$)} was explained in Section \ref{sec:usecase}, which is dependent on the stock value $S_i$ of parcel \textit{i} and the delivery partner used $P_i$. 
FN are parcels predicted as being normal (not lost), which in reality were lost. Hence, the misclassification costs $MC_i$ for parcel \textit{i} of false negative predictions are equal to the parcel's stock value $S_i$. Note that a true negative prediction $TN_i$ or true positive prediction $TP_i$ have no misclassification costs, as the model correctly classify these parcels. $MC_i$ for parcel \textit{i} can be determined by $MC_i = FP \cdot IC(S_i, P_i) + FN \cdot S_i$. The total misclassifcation cost of a decision-making model is computed  
by taking the sum of the misclassification costs of parcels.   






We compare the misclassifcation costs for the developed models with the Business Rules and the scenarios where all parcels are insured or not insured. Table \ref{tab: misclas costs} presents the costs, computed on the 10\% test set. 

\begin{table}[H]
\centering
\begin{tabular}{cccccc}
\hline
\toprule
\textbf{Autoencoder} & \textbf{TN} & \textbf{FP} & \multicolumn{1}{l}{\textbf{FN}} & \multicolumn{1}{l}{\textbf{TP}} & \textbf{  Costs} \\ \midrule
Business rules       & 59706       & 25567       & 125                              & 92                               & \euro116.008,13                             \\ 
Insure nothing          & 85273       & 0           & 218                              & 0                                & \euro92.157,57                              \\ 
Insure all           & 0           & 85273       & 0                                & 218                              & \euro177.299,05                             \\ 
RU-RF                & 77797       & 7476        & 111                              & 106                              & \euro60.948,05                              \\ 
AE-RF              & 74830       & 10443       & 105                              & 112                              & \textbf{\euro55.140,82}                              \\ 
\bottomrule
\end{tabular}
\caption{Total misclassification costs on the 10\% test set (85490 instances) of different decision-making models}
\label{tab: misclas costs}
\end{table}

The misclassification costs of the current business rules are equal to \euro116.008,13. It can be observed that if all parcels are insured the total misclassification costs are significantly higher, i.e. \euro177.299,05. However, if no insurance would be used the total stock value of these parcels equals the misclassification costs, which equals \euro92.157,57. This confirms that the current business rules are too simple.  
The current business rules insure one third of all parcels above 100 euros, while only 0,25\% is actually lost. 
The RU-RF model shows moderately higher misclassification costs (\euro60.948,05). Furthermore, the AE-RF model results in the lowest misclassification costs, which equals \euro55.140,82.    
The difference between the AE-RF model and the current situation is equal to \euro60.867,31 on 10\% of the data. Therefore, the expected decrease in costs is yearly equal to \euro608.673,10 if this model would be used. We can conclude that using machine learning models for insurance decision-making can reduce operational costs significantly. 

\subsection{Insights into process}

Besides the monetary savings, additional process insights could be obtained from the machine learning models using techniques from XAI, among which one of the most commonly used method is SHAP \citep{lundberg2017unified}. SHAP calculates the impact of each feature in the final prediction, which can be used for feature interpretability determination and model explainability. SHAP can be applied to black-box models such as random forests and neural networks. 

\begin{figure}[H]
    \centering
    \includegraphics[width =\textwidth]{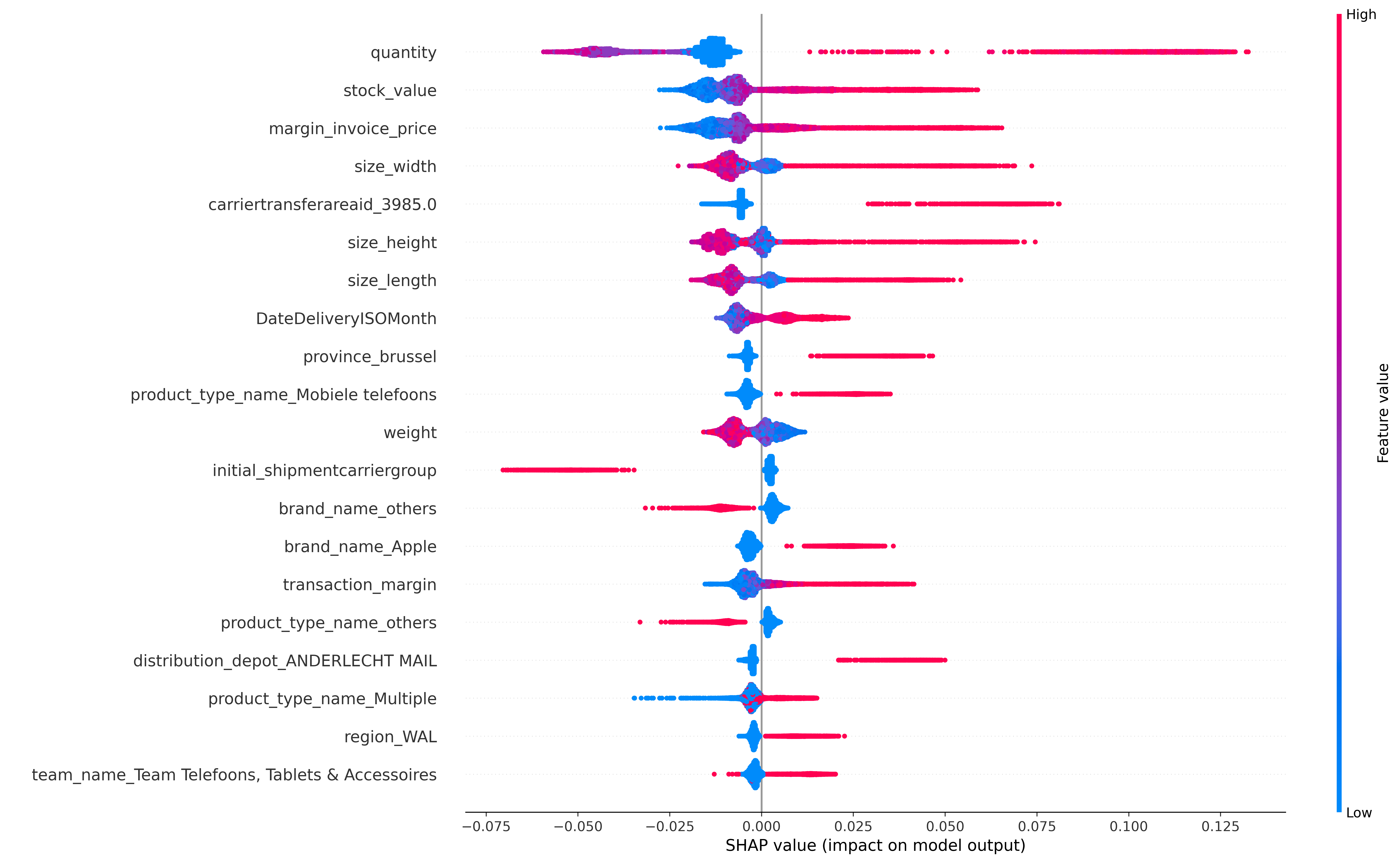}
    \caption{SHAP values for the RU-RF model} 
    \label{fig: SHAP global interpretability}
\end{figure}

However, for the best performing model, i.e., AE-RF, 
we are not aware of any existing techniques from XAI that can be used to interpret the input features, as AE and RF are separately trained on different datasets. More specifically, in AE-RF, an autoencoder adapts the input data, and then the reconstruction errors obtained by the autoencoder serve as inputs to a random forest classifier. Therefore, in the following, we only analyze the interpretability of the developed random forest model, RU-RF. 
SHAP values are obtained from the training dataset and used to establish the feature importances of the random undersampled random forest model. \autoref{fig: SHAP global interpretability} shows the obtained summary plot, which visualizes the feature's importance, impact, and original value. The summary plot lists the 20 most important features of the model obtained during the model's training. The figure is vertically ordered on feature importance, decreasing from top to bottom. The horizontal position on the axis illustrates the impact of a feature. High impact, i.e. high SHAP values, push the prediction towards a positive one and thus parcel loss. The color in the plot provides information regarding the original value of the sample. Red means that the value of the sample is relatively high, while blue indicates a relatively low value of the sample. Combining the impact and the original value provides insight into the correlation of a feature with the target variable. Most interesting are the top features, as these provide the most information during prediction. Generally, the impact and values for features are separated well, which means that the SHAP values provide insight into the relation between parcel loss and feature values. However, it should be noted that the values do not serve as causal relationships but only provide insight into the associations between the features and the target variable learned by the model. 

\autoref{fig: SHAP global interpretability} shows that higher values for the feature quantity push the binary prediction towards a positive one, representing parcel loss. This suggests that multi-product parcels are associated with parcel loss. Similar associations are observed for the stock value and size dimensions, which suggests that high-valued and relatively large parcels are expected to be more often lost. 
Another interesting observation relates to the carrier that is used to deliver parcels. Only OwnDN is listed among the 20 most important features, while ExtD1 and ExtD2 are observed to be less important predictors for parcel loss. It should be noted that this does not mean that there is no relationship between these carriers and parcel loss. The outcomes only give insight into the decision making of the model for final classification. Another remarkable observation is that the delivery method is not listed among the 20 most important features. 

The model seems to associate various features related to the delivery address with parcel loss. The transfer area with ID = 3985, the city of Brussels, the province of Wallonia, and the delivery depot ‘Anderlecht Mail' are associated with more positive parcel loss predictions. 
On the contrary, parcels delivered by OwnDN are associated with less parcel loss. Furthermore, the delivery moment seems to be taken into account in the prediction model. It can be observed that higher values for the created feature DateDeliveryISOMonth are associated with lost parcels. Domain experts state that this is probably due to the high amount of lost parcels during Christmas and Black Friday. Finally, the features representing product attributes are associated with parcel loss. Mobile phones, tablets, and Apple products are often predicted as lost by the model. It seems therefore that a relatively large amount of lost parcels is related to fraud instead of human errors in the process. It should be noted that some of the samples are centered around zero, indicating that the relationship is less strong if the values are not extreme. Overall this leads to the hypothesis that the model's strength lies in the combination of multiple feature values, which suggests that product and order characteristics are combined for prediction. 

\begin{figure}[!ht]
    \subfloat[Feature 'ExtD1'\label{fig:dependence1}]{%
      \includegraphics[width=0.45\textwidth]{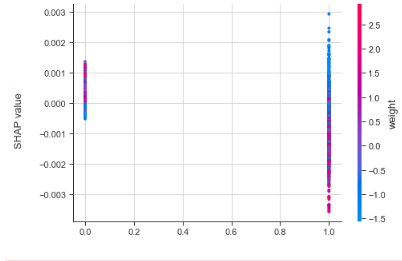}
    }
        \subfloat[Feature  'ExtD2'\label{fig:dependence2}]{%
      \includegraphics[width=0.45\textwidth]{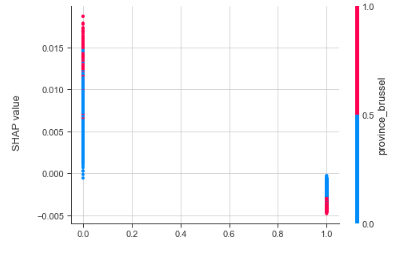}
    }\\
        \subfloat[Feature  'Drop-off'\label{fig:dependence3}]{%
      \includegraphics[width=0.45\textwidth]{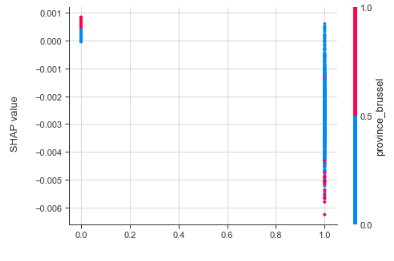}
    } 
    \subfloat[Feature 'Parcel locker'\label{fig:dependence4}]{%
      \includegraphics[width=0.45\textwidth]{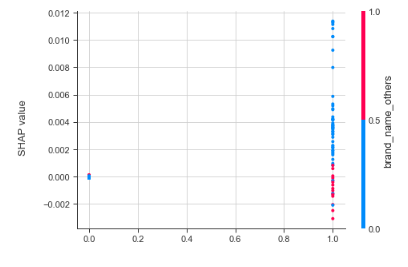}
    }
    \caption{Dependence plots of four binary features.}
    \label{fig:dependence_plots}
  \end{figure}

The existing literature indicates that shipment methods relate to parcel loss. Parcel lockers were expected to be reliable, while drop-off deliveries at pick-up points were prone to parcel loss. Nevertheless, during data exploration it was observed that the absolute amount of deliveries via parcel lockers is low. It is expected that due to the skewed distribution of these features, these binary features are not listed among the 20 most important features and thus primarily ignored. 
In terms of actions that the company can take to prevent parcel loss, product and customer features cannot be altered, while operational decisions such as the carrier selection and the shipment method can be optimized. Therefore, to improve business process, we analyze these features within the prediction model. Figure \ref{fig:dependence_plots} shows the partial dependence plots of the four features related to the shipment method and carrier. The partial dependence plot shows the marginal effect of one feature on the model's predicted outcome. Moreover, this plot shows the feature it interacts most with, depicted by the color. For example, Figure \ref{fig:dependence3} shows solely positive SHAP values (y-axis) for drop-off deliveries with value 0 (x-axis). The value zero in the binary feature indicates that no drop-off is used. This means that the model relates no drop-off with higher probabilities of parcel loss. On the contrary, the value 1 on the x-axis mostly relates to a negative SHAP value on the y-axis. Hence, drop-off deliveries are related to fewer parcel loss. Figure \ref{fig:dependence3}  shows that this observed relationship is strengthened by the binary feature province\_brussel. Drop-off deliveries in Brussels are seen by the model as more reliable deliveries. The three other figures are analysed in the same manner. It can be observed that the carrier ExtD1 seems to have no clear relationship with the predicted outcome of the model. On the other hand, ExtD2 shows a marginal negative effect; ExtD2 deliveries are associated with normal parcel deliveries, while deliveries with other carriers are more susceptible to parcel loss. Finally, it can be observed that parcel lockers show a marginal positive effect. Parcel locker deliveries are expected to become more often lost. These outcomes are very interesting because it conflicts with the hypothesis made in the literature, which suggests that parcel lockers were reliable and drop-off not. 


\subsection{Validation with domain experts}
We discussed the established correlations between the target variable and predictive features with several domain experts at Company X. The stock value is known to be related to parcel loss, as the business rules were fully focused on this feature for insurance decision making. Furthermore, the quantity of products in the parcel is related to the stock value because more products represent a higher stock value. In addition, the relationship between the parcel size is found interesting by domain experts. \autoref{fig: SHAP global interpretability} shows that either relative small or large parcels are related to positive parcel loss predictions. It is expected that fraudsters recognize small parcels from Company X as valuable phones, triggering potential fraud. On the other hand, small parcels are more easily lost due to process errors or could be delivered in the mailbox instead of to the door, which could trigger additional loss. The exact relationship between the parcel size and parcel loss needs to be further analysed by Company X to determine potential process improvements. Large parcels are more easily recognized and fraudsters could expect high valued TV's or computers to be inside, triggering further fraudulent loss. Similar, the product category is known to be related to parcel loss. The current business rules use the product categories: phones, tablets and laptops in the decision making. The model shows that besides these product categories also brands are related to parcel loss, such as Apple. Domain experts state that the delivery depot and location-based features that showed associations to parcel loss were suspected but never empirically supported. Depots seem to be very valuable predictors for parcel loss and can help to improve business processes. Parcel loss was mainly occurred in Brussels. 
Loss prevention specialists at Company X were familiar with this observed problem but could not explain these differences. It is expected that Brussels is prone to fraudulent behavior. The feature Month provided insight into the relationship between time and parcel loss. It was observed that the later months in the year showed relatively more parcel loss. However, as only one year of data is used in this study it is hard to generalize the outcomes. Domain experts state that the observed relation is probably due to the total amount of parcels delivered. During Christmas and Black Friday higher volumes cause more process failures and less overview on processes, which explains the larger amount of observed parcel loss. Finally, it was observed that parcel lockers were susceptible to parcel loss, while drop-off deliveries show the opposite relationship. Domain experts state that drop-off deliveries are more reliable due to the extra person and used proof of delivery upon delivery and pick-up by customers. Moreover, parcel lockers were known to be susceptible to fraud in 2020, as this delivery method was at that moment in time only recently in use. Domain experts state that the expectation is that in the near future parcel lockers will reduce the amount of parcel loss.

\subsection{Managerial implications}
Looking at both the misclassificaton cost and the interpretability, we conclude that the most feasible model is the random under-sampled random forest model (RU-RF) of DBSL, which is expected to save \euro550.600,80 euros yearly for Company X. 

 
Our results confirm that the current business rules of Company X on insurance are outdated and the use of predictive models could save a significant amount of costs. 
Based on our results, we recommend Company X to implement the random under-sampled random forest model (RU-RF) to change insurance decision-making towards a more data-driven method.  As shown in our case study, this implementation will decrease the costs related to parcel loss.  
To further improve the business operations, we suggest that process engineers should analyze the
predictions of the model. We have demonstrated  valuable insights into drivers of parcel loss can be gained from the predictions.  For instance, our analysis show that the quantity, size, and stock value are expected to be positively related to parcel loss, 
and in contrast, drop-off deliveries at pick-up points and deliveries with ExtD2 are expected to be negatively associated with parcel loss. Moreover, parcel lockers are currently likely to be prone to parcel loss. Finally, several product- and delivery address-related features are expected to affect the amount of parcel loss.

In addition, including expected costs of parcel loss in
the carrier selection process can improve processing efficiency and reduce costs related to parcel loss.
For instance, the carrier OwnDN was observed to decrease the amount of parcel loss. Instead of
using insurance on `high-risk' parcels, it should be explored what the effect is of changing the carrier to OwnDN. We expect that this will reduce the amount of parcel loss and increase customer satisfaction.
Furthermore, our results suggest Company X should enhance drop-off deliveries, which are expected to decrease
parcel loss. On the contrary, parcel locker deliveries should be avoided. Finally, Company X is recommended to
do further analyses on parcel packaging and labeling, as our models show that stock value, quantity, and
size are important factors of parcel loss. Fraudsters are expected to recognize high-valued parcels on
appearance, which increases parcel loss.

\section{Conclusion}
\label{sec:conclude}
In this paper, we present a first study on using machine learning to predict parcel losses in the last mile delivery. We did a case study with Company X, one of the largest online retailers in the Netherlands. Our results show that all proposed machine learning methods are able to predict parcel losses much better than the current conventional decision rules used by Company X. In particular,  the proposed deep hybrid model (DHEL) gives the best performance among all tested machine learning approaches, but lacks interpretability. On the other hand, the proposed sampling with supervised learning method (DBSL), especially random undersmapling with random forests, is able to generate useful insights to improve the operational decisions for Company X. 

We want to emphasize that while this study utilizes data from Company X specifically, we believe that the developed Machine Learning methods hold potential for broad applicability across various e-commerce retailers. This is based on the shared characteristics of delivery data within this domain, notably the presence of substantial datasets characterized by imbalanced class distribution.

\bibliographystyle{model5-names}\biboptions{authoryear}

\bibliography{References}

\newpage
\appendix 

\section{Supervised learning methods}
\label{appendix:A}

\textbf{Logistic Regression (LR) }
predicts the probability of an event using a linear combination of features. The logistic function converts linear regression output to a probability between 0 and 1. The threshold is often set at 0.5. LR is simple, interpretable and computationally efficient~\citep{kleinbaum2010assessing}. However, the LR generally performs poorly on high-dimensional data problems. According to \citep{owen2007infinitely}, the LR algorithm is commonly used as
a classification model when one of the two classes is extremely rare, making this model well-suited for this research.

\textbf{Decision Tree (DT)} is a simple machine learning model that uses a tree-like graph to predict outcomes~\citep{breiman2017classification}. It breaks down a data set into smaller and smaller subsets based on the values of decision attributes at each step. The model starts from a root node at the top and splits the data into branches, moving from parent node to child nodes. At each step, the tree chooses the attribute that best splits the data into homogeneous subsets. This repeats until terminal nodes are reached, indicating class labels or regression values. DTs are easy to understand and interpret.

\textbf{Random Forest (RF)} is an ensemble learning model that uses many decision trees during both training and prediction~\citep{breiman2001random}. It constructs many decision trees during training, then outputs the class that most trees predict. This controls overfitting and improves accuracy over a single tree. Each tree uses a random subset of data and predictors, introducing randomness that increases diversity among trees and boosts performance.  By aggregating the predictions from all the trees, the random forest model reaches a robust final prediction. The output from a RF model is based on the average or majority vote of the output from multiple decision trees, making it more accurate and stable.

\textbf{Extreme Gradient Boosting (XGBoost)} is an ensemble of the DT algorithm, where new trees solve the errors of the previous tree model \citep{natekin2013gradient}. Gradient boosting focuses on gradient reduction of the loss function of the previous model. The XGBoost is an improved version of the Gradient Boosting algorithm, which can further mitigate variance. The main improvement is that the loss function is normalized to reduce model variances. XGBoost develops a strong learner by integrating multiple weak learners, reducing the probability of misclassification by iteration. Trees are added until no further improvements can be made to the model.

\textbf{Support Vector Machines (SVM)} plots the  data items into n-dimensional space where $n$ is the number of features~\citep{platt1999probabilistic}. It then determines the decision boundary separating the data into classes. The decision boundary is chosen to maximize the margin between the boundary and the data points of different classes. The data points that lie closest to the decision boundary are called support vectors. These support vectors are important since they define the position of the boundary. For prediction, new data instance is mapped to the space and classified based on its side of the boundary. SVM handles sparsity in high dimensions and can perform nonlinear classification using kernels.
\section{Semi-supervised deep anomaly detection methods}
\label{appendix:B}
\begin{figure}[H]
    \centering
    \includegraphics[width = 0.70\textwidth]{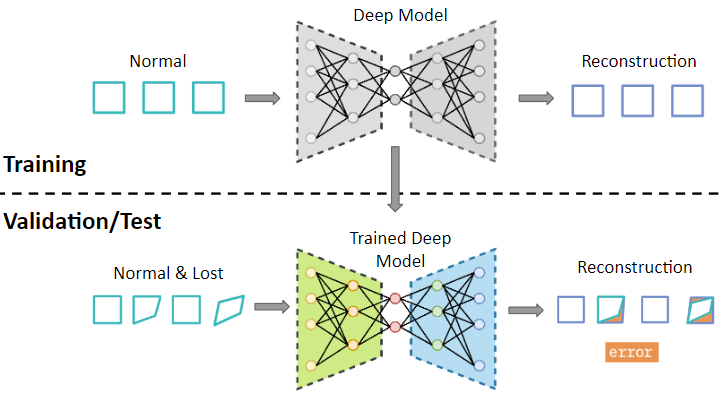}
    \caption{Visualization of semi-supervised deep anomaly detection} 
    \label{fig: semi-supervised learning}
\end{figure}

Anomaly detection is proven to be very successful for binary classification in imbalanced data sets \citep{chalapathy2019deep}. 
Anomaly detection using autoencoders is useful for class imbalance, where autoencoder (AE) is trained only on normal data, for the purpose of learning how to reconstruct this normal data. During training, the AE implicitly learns the important patterns in the normal data that allow it to achieve good reconstruction. At the test time, new data instance is reconstructed resulting in low error for normal data and high error for anomalies. The higher error for anomalies indicates they do not conform to the learned model of normality. By setting a threshold based on the reconstruction error of the normal data, any input with a reconstruction error higher than this threshold can be classified as anomalous. The Figure~\ref{fig: semi-supervised learning} schematically illustrates the training and validation setup of AE. 

\begin{figure}[H]
    \centering
    \includegraphics[width = 0.65\textwidth]{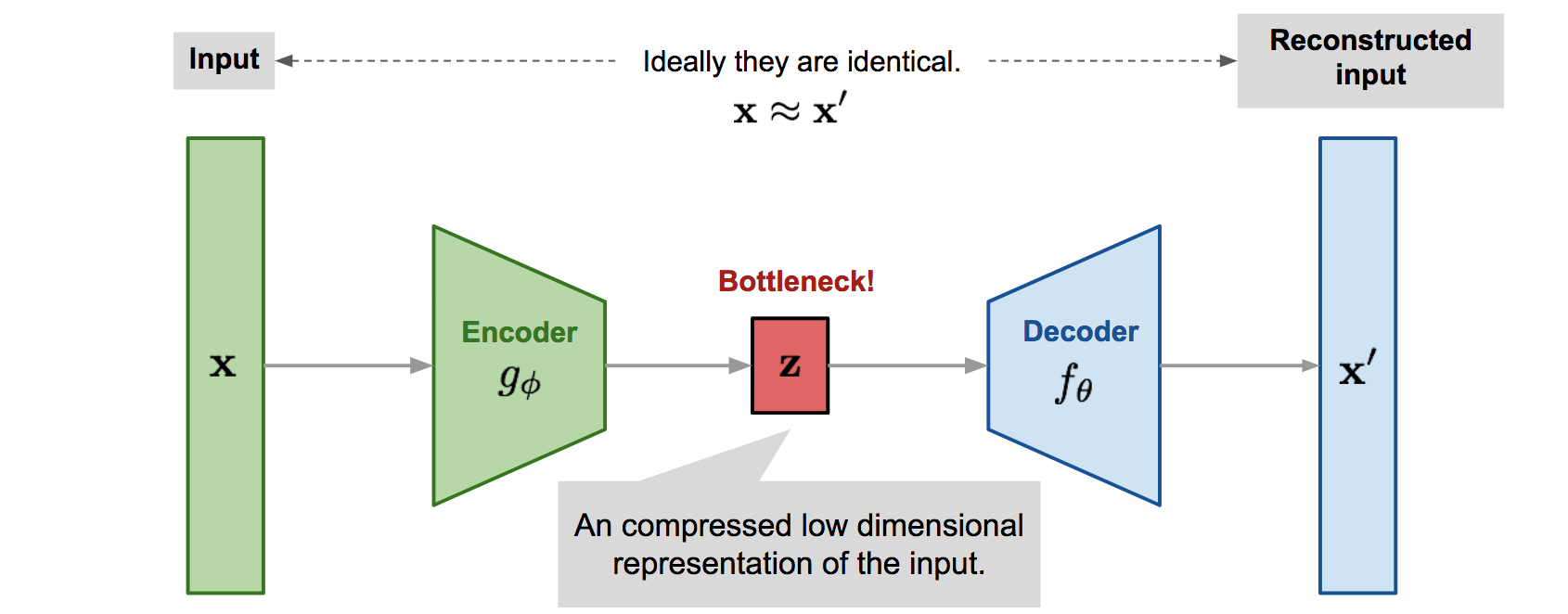}
    \caption{Autoencoder architecture, retrieved from \citep{weng2018VAE}} 
    \label{fig: autoencoder}
\end{figure}

\textbf{Autoencoder (AE)} is an unsupervised neural network that learns to reproduce input vectors by learning low-dimensional feature representations~\citep{hinton2006reducing}. An AE consists of an encoder network and a decoder network. The encoder maps input data into a lower dimensional feature space, referred to as the bottleneck layer or latent representation. The bottleneck layer holds the compressed representation of the input data. The decoder then reconstructs data from the low-dimensional feature space. A schematic illustration of AE  is visualized in \ref{fig: autoencoder}. The encoder function $g(.)$ and decoder function $f(.)$ are parameterized by $\phi$ and $\theta$, respectively. Both of these parameters are learned together to output a reconstructed sample $x'= f_{\theta}(g_\phi(x))$ which should ideally be similar to $x$. 

\begin{figure}[H]
    \centering
    \includegraphics[width = 0.65\textwidth]{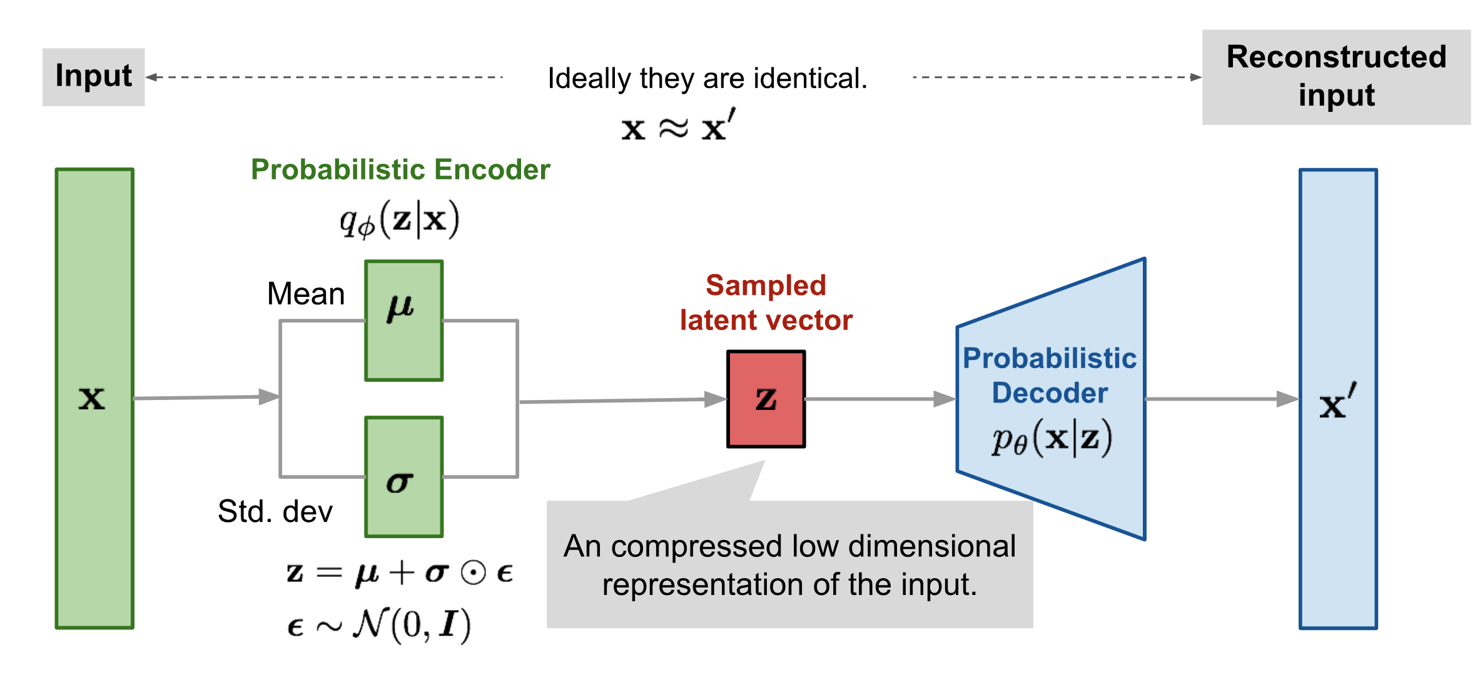}
    \caption{Variational autoencoder architecture, retrieved from \citep{weng2018VAE} } 
    \label{fig: vae2}
\end{figure}

\textbf{Variational Autoencoders (VAE)} is a probabilistic version of the regular AE that can better grasp non-linear data characteristics \citep{kingma2013auto}. The encoder learns the mean $\mu$ and standard deviation $\phi$ of a Gaussian distribution that represents the input data in a continuous latent space. The encoder outputs the mean and standard deviation parameters, representing each data point as a probability distribution in the latent space rather than a deterministic vector. The decoder reconstructs inputs from vectors sampled from this Gaussian distribution. This allows VAEs to model the underlying probability distribution of the dat. The VAE loss function optimizes both reconstruction accuracy and the similarity between the encoded distribution and a prior distribution. A schematic illustration of VAE  is visualized in \ref{fig: vae2}. 

\begin{figure}[H]
    \centering
    \includegraphics[width = 0.65\textwidth]{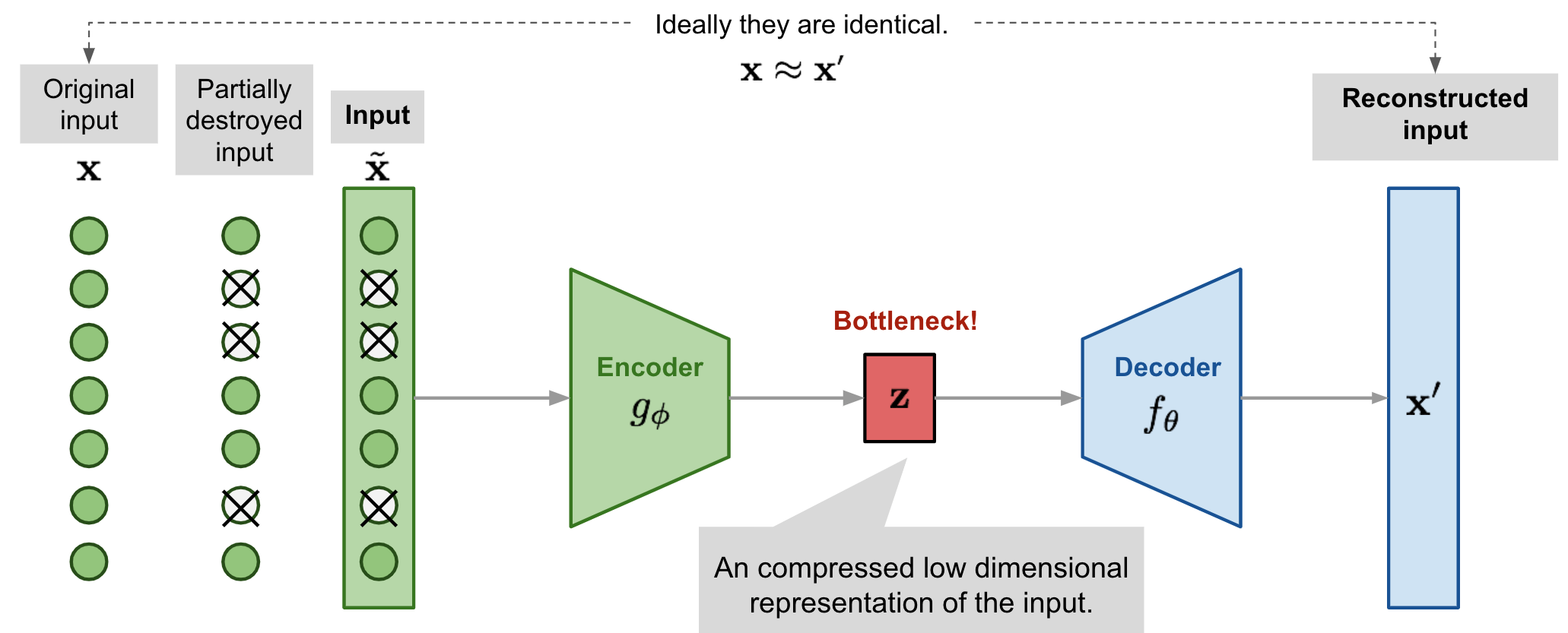}
    \caption{Denoising autoencoder architecture from \citep{weng2018VAE}}
    \label{fig: DAE}
\end{figure}

\textbf{Denoising Autoencoders (DAE)} The standard AE learns the identity function which can lead to overfitting. The DAE model proposes to to corrupt the input data by adding stochastic noise, $\tilde{\mathbf{x}} \sim \mathcal{M}_\mathcal{D}(\tilde{\mathbf{x}} \vert \mathbf{x})$, where $\mathcal{M_D}$ defines the mapping from true data samples to the noisy samples~\citep{vincent2008extracting}. The model is then trained to recover $x'$ from the input $x$ without the noise. The amount of noise that results in the best performing model is problem dependent and is thus an additional hyperparameter $\sigma_n$, which is optimized in this research. A schematic illustration of AE  is visualized in \autoref{fig: DAE}. 

\end{document}